\documentclass{sigchi}

\CopyrightYear{2019}
\setcopyright{acmlicensed}
\doi{https://doi.org/10.1145/3332165.3347911}
\isbn{978-1-4503-6816-2/19/10}
\conferenceinfo{UIST'19,}{October 20--23, 2019, New Orleans, LA, USA}
\acmPrice{\$15.00}

\usepackage{balance}
\usepackage{graphics}
\usepackage[T1]{fontenc}
\usepackage{txfonts}
\usepackage{mathptmx}
\usepackage[pdflang={en-US},pdftex]{hyperref}
\usepackage{color}
\usepackage{booktabs}
\usepackage{textcomp}
\usepackage{microtype}
\usepackage{ccicons}
\usepackage{todonotes}
\usepackage{listings}

\def\plaintitle{ShapeBots: Shape-changing Swarm Robots}

\def\emptyauthor{}
\def\plainkeywords{shape-changing user interfaces, swarm user interfaces, tangible interactions;}

\makeatletter
\def\url@leostyle{%
  \@ifundefined{selectfont}{
    \def\UrlFont{\sf}
  }{
    \def\UrlFont{\small\bf\ttfamily}
  }}
\makeatother
\def\pprw{8.5in}
\def\pprh{11in}

\definecolor{linkColor}{RGB}{6,125,233}
\hypersetup{%
  pdftitle={\plaintitle},
  pdfauthor={\emptyauthor},
  pdfkeywords={\plainkeywords},
  pdfdisplaydoctitle=true, 
  bookmarksnumbered,
  pdfstartview={FitH},
  colorlinks,
  citecolor=black,
  filecolor=black,
  linkcolor=black,
  urlcolor=linkColor,
  breaklinks=true,
  hypertexnames=false
}

\usepackage{enumitem}

\newcommand {\changes}[1]{{#1}}

\begin{document}
\title{
ShapeBots: Shape-changing Swarm Robots
\vspace{-0.5cm}
}

\numberofauthors{1}
\author{
  \alignauthor{
    Ryo Suzuki$^1$,
    Clement Zheng$^2$,
    Yasuaki Kakehi$^3$,
    Tom Yeh$^1$,
    Ellen Yi-Luen Do$^{1,2}$\\
    Mark D. Gross$^{1,2}$,
    Daniel Leithinger$^{1,2}$\\
  \affaddr{
    $^1$Department of Computer Science, University of Colorado Boulder,\\
    $^2$ATLAS Institute, University of Colorado Boulder,
    $^3$The University of Tokyo
  }\\
  \email{
    \{ryo.suzuki, clement.zheng, tom.yeh, ellen.do, mdgross, daniel.leithinger\}@colorado.edu, kakehi@iii.u-tokyo.ac.jp
  }}\\
}

\teaser{
\vspace{-0.7cm}
\centering
\includegraphics[width=1\textwidth]{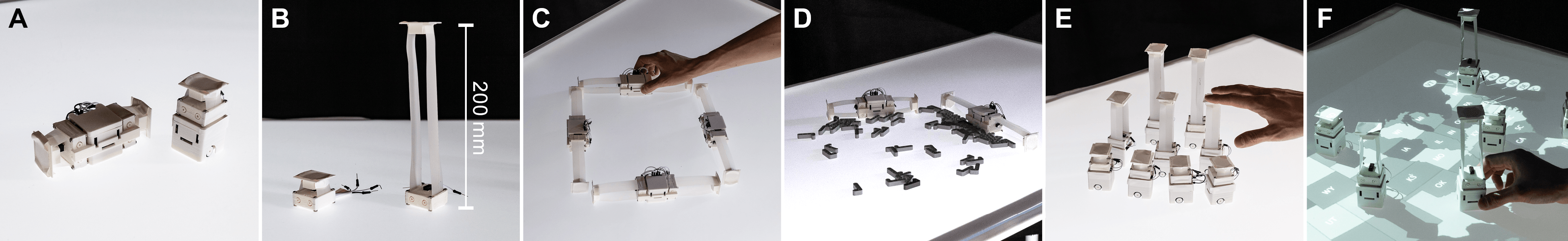}
\caption{
ShapeBots exemplifies a new type of shape-changing interface that consists of a swarm of self-transformable robots. A) Two ShapeBot elements. B) A miniature reel-based linear actuator for self-transformation. By leveraging individual and collective transformation, ShapeBots can provide C) interactive physical display (e.g., rendering a rectangle), D) object actuation (e.g., cleaning up a desk), E) distributed shape display (e.g., rendering a dynamic surface), and F) embedded data physicalization (e.g., showing populations of states on a US map).
}
~\label{fig:cover}
\vspace{-1.1cm}
}

\maketitle

\begin{abstract}
We introduce {\it shape-changing swarm robots}. A swarm of self-transformable robots can both {\it individually} and {\it collectively} change their configuration to display information, actuate objects, act as tangible controllers, visualize data, and provide physical affordances. ShapeBots is a concept prototype of shape-changing swarm robots. Each robot can change its shape by leveraging small linear actuators that are thin (2.5 cm) and highly extendable (up to 20cm) in both horizontal and vertical directions. The modular design of each actuator enables various shapes and geometries of self-transformation. We illustrate potential application scenarios and discuss how this type of interface opens up possibilities for the future of ubiquitous and distributed shape-changing interfaces.
\end{abstract}

\vspace{-0.3cm}


\vspace{-0.3cm}


\vspace{-0.3cm}

\section{Introduction}
Today, we live in a world where computers and graphical displays almost ``disappear'' and ``weave into the fabric of everyday life''~\cite{weiser1991computer}. Dynamic graphical interfaces---e.g., smartphones, smartwatches, projectors, and digital signage---are now distributed and embedded into our environment. We envision dynamic physical interfaces---e.g., actuated tangible interfaces~\cite{poupyrev2007actuation}, robotic graphics~\cite{mcneely1993robotic}, and shape-changing interfaces~\cite{coelho2011shape, rasmussen2012shape}---will follow the same path as technology advances. Although current interfaces are often large, heavy, and immobile, these interfaces will surely be replaced with hundreds of distributed interfaces, in the same way that desktop computers were replaced by hundreds of distributed mobile computers. If shape-changing interfaces will become truly ubiquitous, how can these interfaces be distributed and embedded into our everyday environment?

This paper introduces {\it shape-changing swarm robots} for distributed shape-changing interfaces.
Shape-changing swarm robots can both {\it collectively} and {\it individually} change their shape, so that they can collectively present information, act as controllers, actuate objects, represent data, and provide dynamic physical affordances. 

Shape-changing swarm robots are inspired by and built upon existing swarm user interfaces~\cite{braley2018griddrones, kim2017ubiswarm, le2016zooids, le2019dynamic, suzuki2018reactile}. Swarm user interfaces support interaction through the collective behaviors of many movable robots. 
By combining such capability with individual shape change, we can enhance the expressiveness, interactions, and affordances of current swarm user interfaces. 
For example, self-transformable swarm robots can support representations that are not limited to moving points, but also lines, and other shapes on a 2D surface. 
For example, each little robot could change its width or height to display a geometric shape (Figure~\ref{fig:cover}C) or represent data embedded in the physical world (Figure~\ref{fig:cover}F). By collectively changing their heights, they can also render a dynamic shape-changing surface (Figure~\ref{fig:cover}E).
In addition to rendering information, it can enhance interactions and affordances of current shape-changing interfaces (Figure~\ref{fig:cover}D). For example, these robots can collectively behave to actuate existing objects (e.g., clean up a desk, bringing tools when needed), become a physical constraint (e.g., a shape-changing ruler), provide physical affordances (e.g., create a vertical fence to indicate that a coffee cup is too hot to touch), and serve as a tangible controller (e.g., for a pinball game).

We developed ShapeBots, self-transformable swarm robots with modular linear actuators (Figure~\ref{fig:cover}A). 
One technical challenge to enable shape-changing swarm robots is the development of small actuators with a large deformation capability. 
To address this, we developed a miniature reel-based linear actuator that is thin (2.5 cm) and fits into the small footprint (3 cm x 3 cm), while able to extend up to 20 cm in both horizontal and vertical directions (Figure~\ref{fig:cover}B).
The modular design of each linear actuator unit enables the construction of various shapes and geometries of individual shape transformation (e.g., horizontal lines, vertical lines, curved lines, 2D area expansion, and 3D volumetric change).
Based on these capabilities, we demonstrate application scenarios showing how a swarm of distributed self-transformable robots can support everyday interactions.

Beyond the specific implementation of the ShapeBots prototype, we outline a broader design space for {\it shape-changing swarm user interfaces}.
We discuss future research opportunities towards ubiquitous shape-changing interfaces.

\begin{figure}[!t]
\vspace{-0.2cm}
\centering
\includegraphics[width=3.4in]{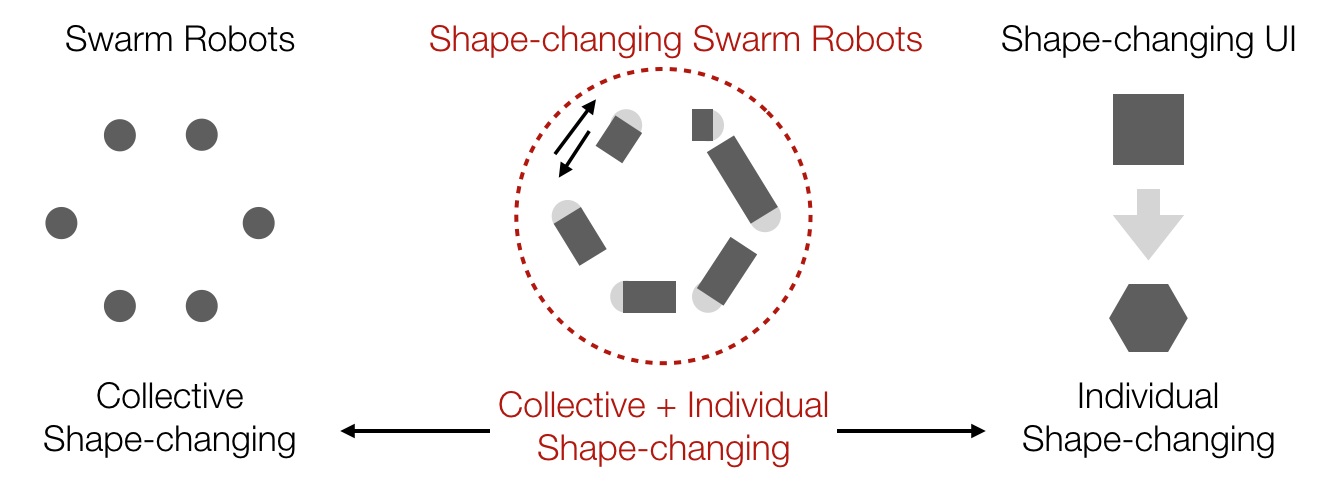}
\caption{Swarm robots leverage collective shape transformation (left), and shape-changing interfaces leverage an individual shape transformation (right). Shape-changing swarm robots leverage both \textit{collective} and \textit{individual} shape transformation (center).}
~\label{fig:concept-diagram}
\vspace{-0.8cm}
\end{figure}

In summary, this paper contributes:
\begin{enumerate}[itemsep=-1mm]
\item A concept of shape-changing swarm robots, a distributed shape-changing interface that consists of a swarm of self-transformable robots.
\item ShapeBots implementation\footnote{\url{https://github.com/ryosuzuki/shapebots}} with a novel linear actuator that achieves a high extension ratio and small form factor.
\item A set of application scenarios that illustrate how ShapeBots can enhance the display, interactions, and affordances of the current swarm and shape-changing interfaces. 
\item Design space exploration of shape-changing swarm user interfaces and discussion for future research opportunities.
\end{enumerate}


\section{Related Work}

\subsection{Swarm User Interfaces}
Swarm user interfaces (Swarm UIs) are a class of computer interfaces that leverage many collective movable physical elements (e.g., 10-30) for tangible interactions~\cite{le2016zooids}.
Swarm user interfaces have several advantages over existing shape-changing interfaces. 
For example, swarm UIs are not constrained in a specific place as they can move freely on a surface~\cite{le2016zooids, suzuki2018reactile}, on a wall~\cite{kim2017ubiswarm}, on body~\cite{dementyev2016rovables}, or even in mid-air~\cite{braley2018griddrones}.
In addition, a swarm of robots provides scalability in shape change as it comprises many interchangeable elements.
The number of elements can also be flexibly changed which contributes to both scalability and expressiveness of displaying information.
Swarm UIs transform their overall shape by collectively rearranging individual, usually identical units. The ability to heterogeneously transform individual shapes can expand the range of expressions, interactions, and affordances. This can be useful in many fields that current swarm UIs support, such as geometric expressions~\cite{le2016zooids}, iconic shapes and animations~\cite{kim2017ubiswarm} (e.g., animated arrow shape), education and scientific visualizations~\cite{ozgur2017cellulo, suzuki2018reactile} (e.g., visualization of magnetic fields or sine waves), physical data representations~\cite{le2019dynamic} (e.g., line graphs, bar graphs, network graphs), accessibility~\cite{guinness2018haptic, suzuki2017fluxmarker} (e.g., tactile maps), and tangible UI elements~\cite{patten2007mechanical, patten2014thumbles} (e.g., a controller and a slider). Thus, this paper presents how the additional capability of self-transformation can augment current swarm UIs.

\subsection{Line-based Shape-Changing Interfaces}
Our work is also inspired by line-based shape-changing interfaces~\cite{nakagaki2017designing}. 
Recent work has shown the potential of an actuated line to represent various physical shapes and to provide rich affordances through physical constraints.
For example, LineFORM~\cite{nakagaki2015lineform} demonstrates how a physical line made from actuated linkages can transform into a wristwatch, a phone, and a ruler to afford different functionality.
Moreover, highly extendable linear actuators can achieve both shape- and size-changing transformations~\cite{hammond2017pneumatic, hawkes2017soft, takei2012morphys}.
G-Raff~\cite{kim2015g} and HATs~\cite{mi2011hats} have explored height-changing tangible interfaces. These interfaces can synchronize the height of objects with digital content on a tabletop surface.
Our work extends such line-based shape-changing interfaces, where each line can not only deform but also move and collectively form a shape for tangible interactions.

\subsection{Modular Shape-Changing Interfaces}
Modular shape-changing interfaces promise to increase flexibility and scalability of design and shape changes. For example, Topobo~\cite{raffle2004topobo} and ShapeClip~\cite{hardy2015shapeclip} allow a designer to construct different geometries of shape-changing interfaces. Changibles~\cite{roudaut2014changibles} and Cubimorph~\cite{roudaut2016cubimorph} are shape-changing robots that leverage a modular and reconfigurable design to achieve different geometries. ChainFORM~\cite{nakagaki2016chainform} integrates modular sensing, display, and actuation to enhance interactions. However, most of these interfaces have limited locomotion.
Outside of HCI contexts, modular reconfigurable robotics~\cite{yim2007modular}, which leverages self-assembly for shape transformation is an active research area. Although this approach promises an ultimate shape-changing interface, with arbitrary shape transformation~\cite{goldstein2005programmable, sutherland1965ultimate}, transformation is often too slow for real-time user interactions. Therefore, a key consideration for our self-transformable robots is fast locomotion and transformation, which our prototype demonstrates.


\section{Shape-changing Swarm User Interfaces}

\subsection{Definition and Scope}
\begin{figure}[!h]
\centering
\includegraphics[width=3.4in]{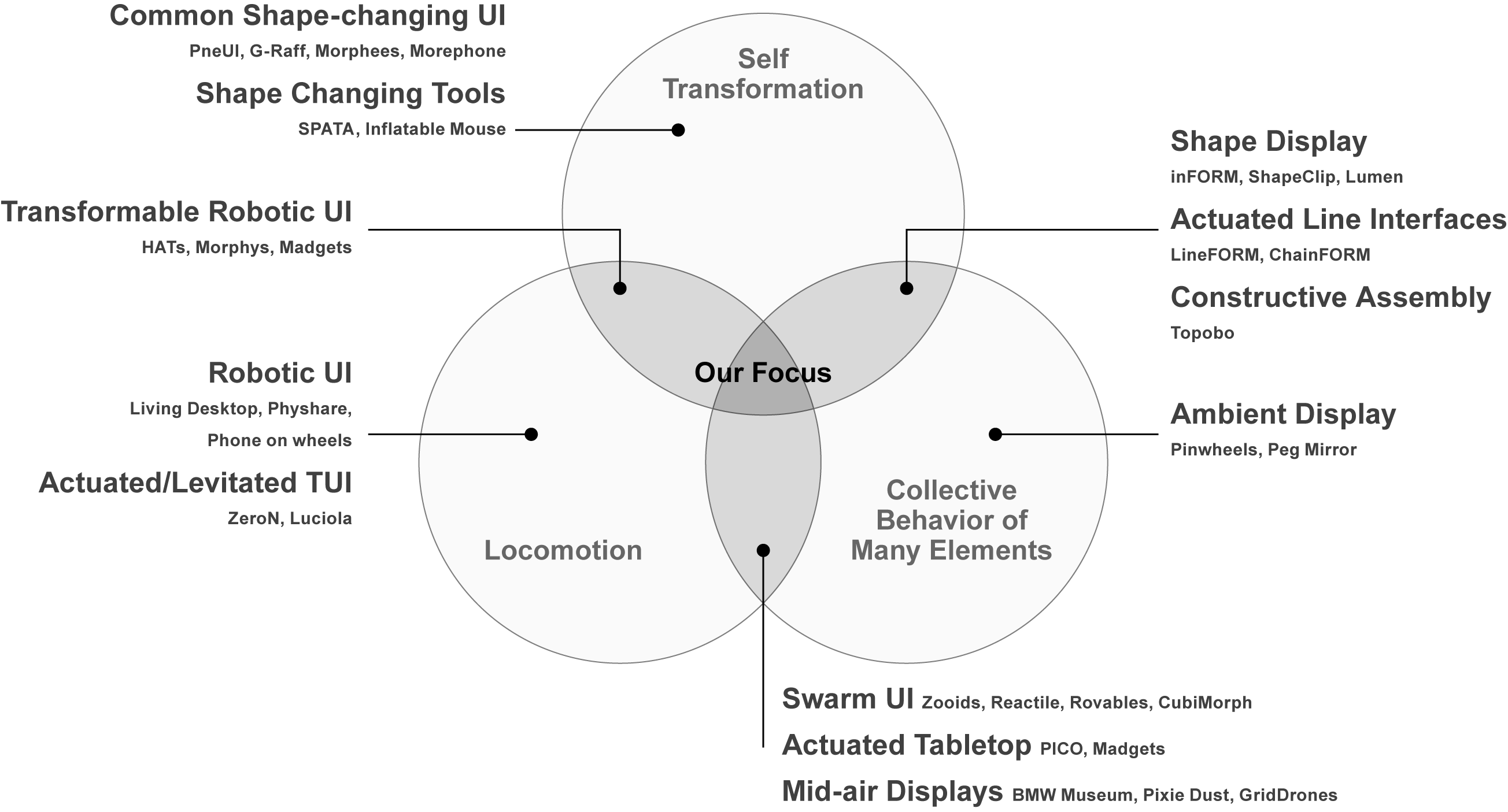}
\caption{Scope and definition of \textit{shape-changing swarm user interfaces}. The diagram classifies and highlights the difference between existing shape-changing interface systems and shape-changing swarm UIs.}
~\label{fig:definition}
\vspace{-0.6cm}
\end{figure}

We introduce shape-changing swarm robots as a type of system that consists of a swarm of self-transformable and collectively movable robots. This paper specifically focuses on the user interface aspect of such systems, which we refer to {\it shape-changing swarm user interfaces}. We identified three core aspects of shape-changing swarm robots: 1) locomotion, 2) self-transformation, and 3) collective behaviors of many individual elements (Figure~\ref{fig:definition}).

\subsubsection{Locomotion}
Locomotion is the key to make shape-changing interfaces more ubiquitous, which gives an interface more agency to be at hand on demand. For example, the device can come to the user and provide in-situ help when needed,  then disappear when no longer needed. Locomotion also expands the space where the user can interact. For example, shape-changing interfaces that can move on a table, walls, and bodies can free the interaction space from a fixed specific location, as mobile and wearable devices have done.

\subsubsection{Self-transformation}
Self-transformation refers to shape changes of a single element (e.g., form, texture, volume), as opposed to shape changes through spatial distributions.
The shape of a physical object affords use and functionality. Thus, the capability of self-transformation plays an important role to provide rich physical affordances for tangible interactions. 

\subsubsection{Collective Behaviors of Many Elements}
A single shape-changing object is limited in its ability to represent general shapes.
Collective behaviors of many elements can overcome this limitation. Many actuated elements can act together to create expressive and general-purpose shape transformation that a single actuated element cannot achieve. Shape displays leverage this aspect to present arbitrary 2.5D shapes on a tabletop surface. 

Given these core aspects, we categorized current shape-changing user interface research in Figure~\ref{fig:definition} Our focus on {\it shape-changing swarm user interfaces} lies in the intersection of these three aspects. For example, shape displays leverage both the self-transformation of each actuator and the collective behaviors of many elements. In contrast, swarm user interfaces and actuated tabletop leverage locomotion and many collective elements. Transformable robotic interfaces leverage self-transformation and locomotion. {\it Shape-changing swarm user interfaces} exhibit self-transformation and locomotion, and leverage the collective behavior of many individual elements.

\section{ShapeBots}
ShapeBots are a swarm of self-transformable robots that demonstrate the concept of shape-changing swarm user interfaces. ShapeBots have four key technical components: 1) a miniature reel-based linear actuator, 2) self-transformable swarm robots, 3) a tracking mechanism, and 4) a control system. 

\subsection{Miniature Reel-based Linear Actuators}
To determine the form factor of the shape-changing swarm robot, we adopted a line-based approach. As described above, line-based structures have been used in several existing shape-changing interfaces. 
An advantage of this approach is its flexibility of transformation.
While each unit has a simple straight or curved line structure, combining multiple lines in different directions enables a higher degree of freedom of shape transformation. 
Also, attaching an expandable enclosure to these lines (e.g., an origami sheet or an expandable sphere) allows different geometrical shapes (area and volumetric change).
Therefore, we designed a modular linear actuation unit that we combine in each swarm robot.

\subsubsection{Mechanical Design}
One technical challenge to realize self-transformable swarm robots is the design of a miniature actuator that fits into a small robot and has a large deformation capability. Typical linear actuators, such as a lead screw or a rack and pinion, have only small displacement; they can extend between two to four times in length. One of our main technical contributions is the design of a miniature linear actuator that extends from 2.5 cm to 20 cm (Figure~\ref{fig:cover}B).

\begin{figure}[!h]
\vspace{-0.2cm}
\centering
\includegraphics[width=3.4in]{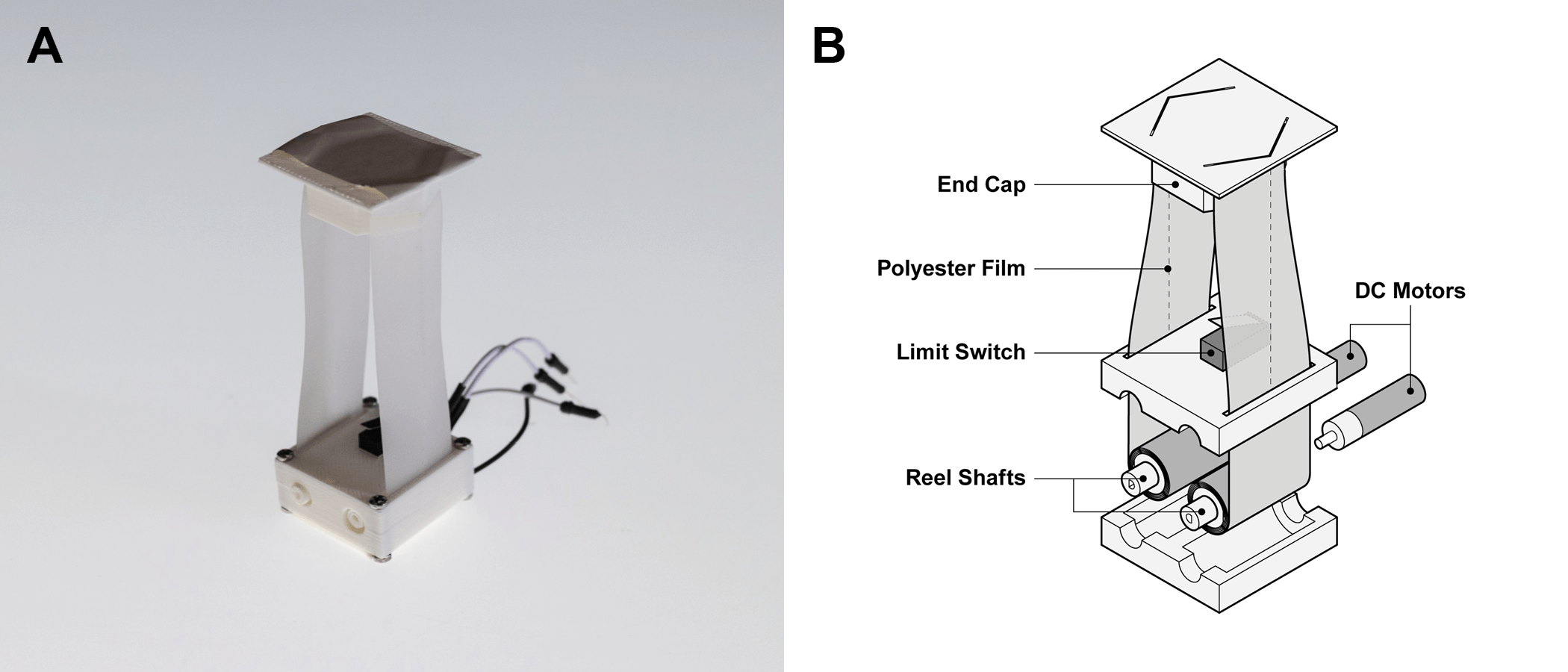}
\caption{The ShapeBot's linear actuation unit, with two micro DC motors, a 3D printed enclosure, and two polyester sheets attached to rotating shafts. By driving these motors, sheets can extend and retract like a tape measure.}
~\label{fig:system-linear-actuator}
\vspace{-0.6cm}
\end{figure}

Figure~\ref{fig:system-linear-actuator} illustrates the design of our linear actuator. It is inspired by a retractable tape measure, which occupies a small footprint, but extends and holds its shape while resisting loads along certain axes. Our reel-based linear actuator employs small DC motors (TTMotor TGPP06D-700, torque: 900g/cm, diameter: 6 mm, length: 22 mm). The linear actuator comprises two reels of thin sheets (e.g. 0.1mm thick polyester sheet). Two DC motors (TTMotor TGPP06D-700) rotate in opposite directions to feed and retract the sheets. Each sheet is creased at the center along its length, and is connected to a cover cap that maintains the fold. This crease increases the structural stability of the linear actuator, similar to how a tape measure remains stable when extended. Thus the linear actuator can push and drag lightweight objects without bending and extend vertically without buckling.

The linear actuator extends and retracts with open-loop control; we estimate extension based on the duration of operating the motors. When fully retracted, the cap triggers a limit switch (Figure~\ref{fig:system-linear-actuator}B), then the length of the actuator will be initialized to zero. The average difference between the target and actual length is less than 5 mm (see the technical evaluation section).
The polyester sheets are attached to 3D printed shafts with super glue. The 3D printed enclosure measures 3cm x 3cm and is 1.5 cm thick. There is 1 cm offset between the cap and the bottom end for the limit switch. Thus, the initial thickness of the linear actuator is 2.5 cm. Although the length of the reel can exceed 20 cm to achieve a higher extension ratio, such a longer reel would more likely buckle.
During prototyping, we observed that friction between sheets and enclosures can cause a jam while extending. Thus, reducing friction is the key to reliable actuation. To reduce friction, we attached a smooth material sheet (e.g., the polyester sheet or peeling sheet of double-sided tape) to the inside of the enclosure.

\subsection{Self-Transformable Swarm Robot}

\subsubsection{Mechanical Design}
\begin{figure}[!h]
\vspace{-0.2cm}
\centering
\includegraphics[width=3.4in]{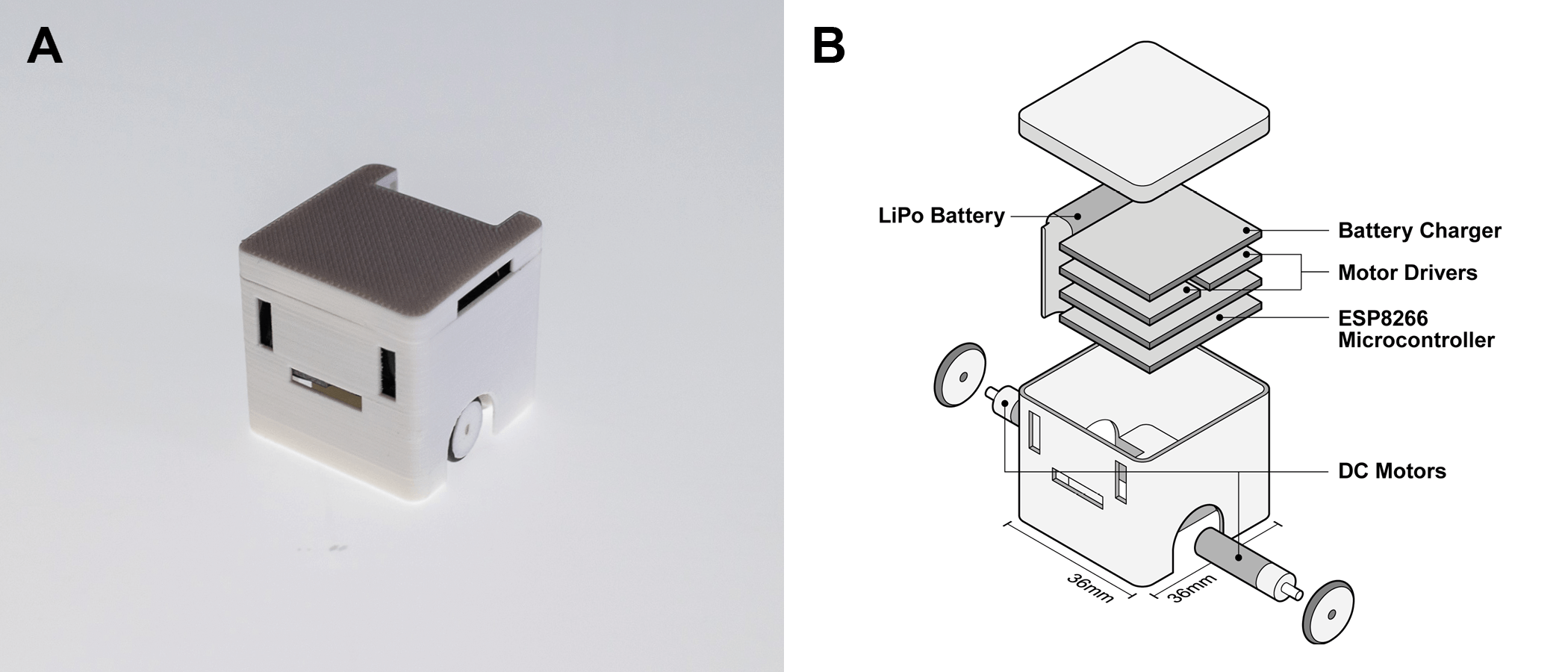}
\caption{The ShapeBot's swarm robot unit. Each swarm robot is driven by two DC motors and controlled through wirelessly. Three rectangular holes in the front face access a programming port, recharging port, and the microcontroller reset switch.}
~\label{fig:system-robot}
\vspace{-0.6cm}
\end{figure}

Figure~\ref{fig:system-robot} illustrates the design of the swarm robot. Each robot is driven by two micro DC motors (TTMotor TGPP06D-136, torque: 550 g/cm, diameter: 6 mm, length: 18 mm). By individually controlling rotation speed and direction, the robot moves forward and backward and turns left and right. Two 3D printed wheels (1 cm diameter) connect directly to the DC motors. An O-ring tire on each wheel increases friction with the ground to avoid slipping. 

Two DC motors soldered to the dual motor driver (DRV8833) are controlled by the main microcontroller (ESP8266). A LiPo battery (3.7V 110mAh) powers both the microcontroller and the motors. Each robot also has an additional DRV8833 motor driver to control the linear actuators; the two motor drivers connect to the microcontroller through a 2-sided custom PCB. All components are enclosed with a 3D printed housing (3.6 cm x 3.6 cm x 3 cm) with three rectangular holes in the front side (Figure~\ref{fig:system-robot}) that house micro USB ports for programming and recharging and the microcontroller reset switch. All 3D printed parts are fabricated with an FDM 3D printer (Cetus 3D MKII) and PLA filament (Polymaker PolyLite 1.75mm True White).
In our prototype, one swarm robot costs approximately 20-25 USD (microcontroller: 4 USD, motor drivers: 3.5 USD x2, DC motors: 3 USD x2, charger module: 1 USD, LiPo battery: 4 USD, PCB: 1 USD) and each linear actuator costs approximately 6-7 USD (DC motors: 3 USD x2, limit switch: 0.5 USD, polyester sheet: 0.1 USD), but this cost can be reduced with volume. For our prototype, we fabricated thirty linear actuator units for twelve robots.

\begin{figure}[!h]
\vspace{-0.2cm}
\centering
\includegraphics[width=3.4in]{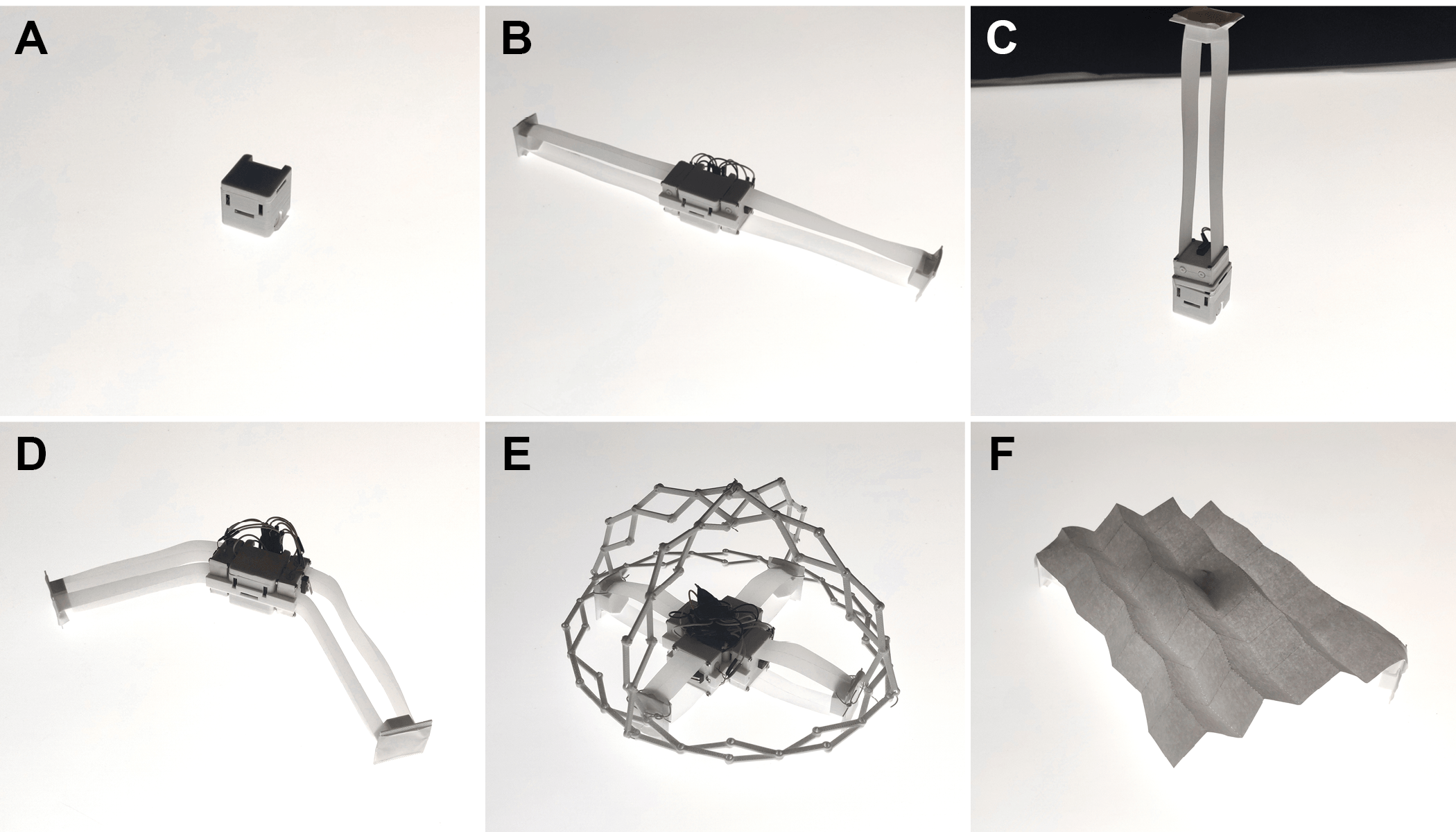}
\caption{Different types of transformation enabled by modular linear actuator units. A) the basic ShapeBot, B) horizontal extension, C) vertical extension, D) bending, E) volume expansion, and F) area expansion.}
~\label{fig:system-transformations}
\vspace{-0.6cm}
\end{figure}

\subsubsection{Types of Transformation}
Due to the modular and reconfigurable design of the linear actuator unit, ShapeBots can achieve several different types of transformations. Figure~\ref{fig:system-transformations} demonstrates five types of shape transformations: horizontal, vertical, and "curved" lines, volumetric change with an expandable Hoberman sphere, and 2D area coverage with an expandable origami structure. These configurations support three types of shape change (e.g., form, volume, orientation) categorized in Rasmussen et al.~\cite{rasmussen2012shape}. 
For horizontal extension, each linear actuator unit is fixed with a custom 3D printed holders. For the vertical extension, we used a thick double-sided tape (3M Scotch Mounting Tape 0.5 inch) on top of the swarm robot.

\subsubsection{Electrical Design}
\begin{figure}[!h]
\vspace{-0.2cm}
\centering
\includegraphics[width=2.6in]{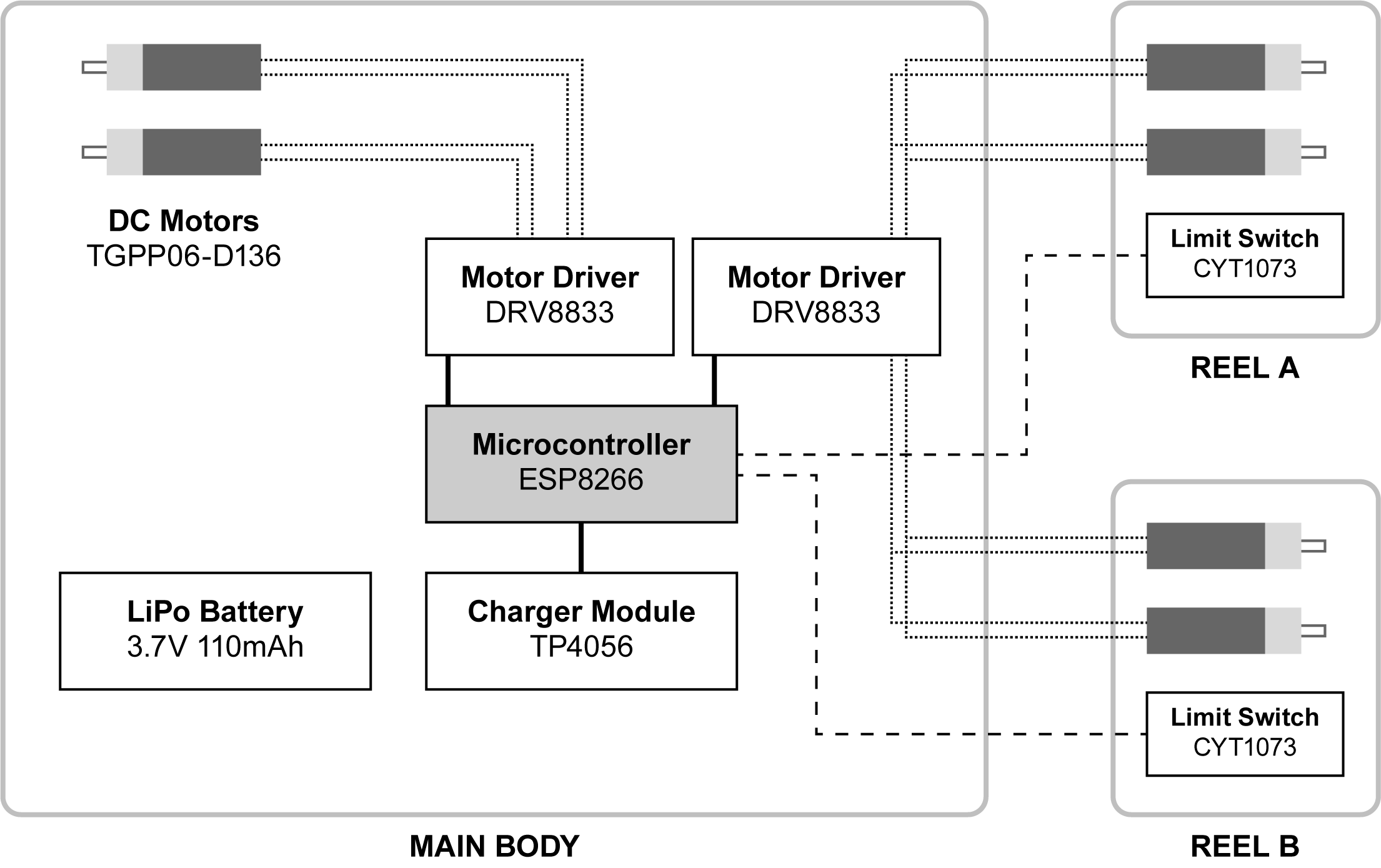}
\caption{Schematics of ShapeBot's electronic components. The main ESP8266 microcontroller operates at 3.3V. Two dual motor drivers drive DC motors for the robot and linear actuators respectively.}
~\label{fig:system-schematics}
\vspace{-0.6cm}
\end{figure}

Figure~\ref{fig:system-schematics} illustrates the schematic of ShapeBots' electronic components. The microcontroller controls two motor drivers, one to operate the robot's two wheel motors, and another to extend the linear actuator.
A single motor driver can only control two actuators independently, but we can actuate more than two actuators by combining signal pins, although this configuration loses the ability to individually control different actuators. With synchronous control, a single robot can operate four linear actuators. In this way, we achieved volume expansion, which requires more than two linear actuators. 

The microcontroller (ESP8266 module) is controlled by the main computer (iMac, Intel 3.2GHz Core i5, 24GB memory) over Wi-Fi. On boot-up, each module connects to a private network and is assigned a unique IP address. The computer sends control commands to each IP address through a user datagram protocol (UDP), namely PWM (0-1023) values for driving wheels and the target length of the linear actuator. These commands control the direction and speed of the swarm robot and the extension of the linear actuators. 

\subsection{Tracking Mechanism}

\subsubsection{Fiducial Markers and Computer Vision Tracking}
To track the position and orientation of the swarm robots, we used computer vision and a fiducial marker attached to the bottom of the robot. Precise orientation tracking is critical. For example, to make a triangle with horizontal extended lines, three swarm robots must orient to appropriate directions. The fiducial marker enables easy, precise, and reliable tracking of both position and orientation that is undisturbed when users occlude the sides and top of the robot during the interaction. Also, it is difficult to track fiducials on top of the robot if it has a vertically extending linear actuator or is covered with an expandable enclosure. We thus decided on tracking the robot from the bottom of a transparent tabletop.

\begin{figure}[!h]
\vspace{-0.2cm}
\centering
\includegraphics[width=3.4in]{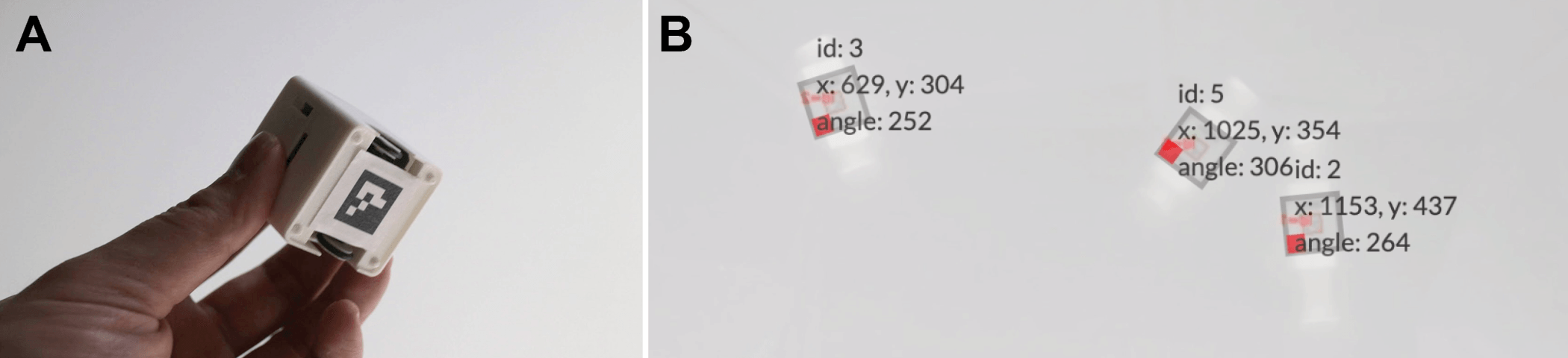}
\caption{A) Fiducial marker (Aruco 4 x 4 pattern, 1.5cm x 1.5cm) is attached to the bottom of each robot. B) OpenCV tracks positions and orientations of markers at 60 FPS.}
~\label{fig:system-fiducial-marker}
\vspace{-0.6cm}
\end{figure}

We used the ArUco fiducial marker~\cite{garrido2014automatic} printed on a sheet of paper and taped to the bottom of the robot.
Our prototype used a 1.5 cm x 1.5 cm size marker with a 4 x 4 grid pattern, which can provide up to 50 unique patterns. For tracking software, we used the OpenCV library and ArUco python module. It can track the position of the markers at 60 frames per second. The captured image and position information is streamed to a web user interface through a web socket protocol. Figure~\ref{fig:system-fiducial-marker} shows the fiducial marker and the captured image.

\begin{figure}[!h]
\vspace{-0.2cm}
\centering
\includegraphics[width=2.6in]{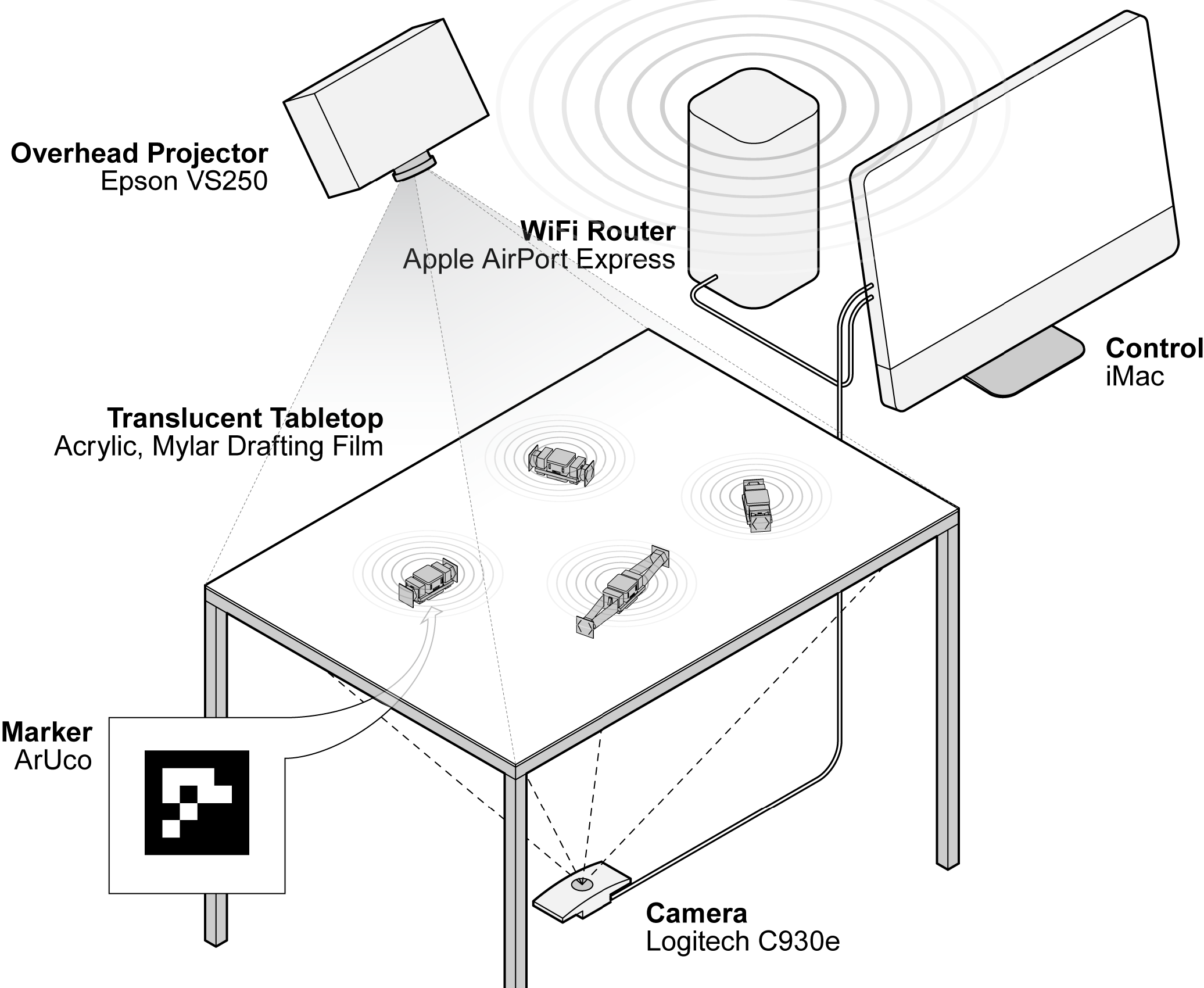}
\caption{A webcam (Logitech C930e) mounted 90 cm beneath the table and connected to the main computer (iMac) captures 115 cm x 74 cm effective area.}
~\label{fig:system-tracking-setup}
\vspace{-0.6cm}
\end{figure}

\subsubsection{Tracking Setup}
We constructed a transparent table with a 6mm acrylic plate mounted on a custom frame. The web camera (Logitech C930e) beneath the table captures a 115 cm x 74 cm effective area. The camera is connected to the main computer that uses the tracking information to control the robots.
We cover the transparent plate with a polyester sheet (Mylar drafting film) to diffuse light and reduce hot spots, which cause unreliable tracking. Three lamps illuminate the bottom of the table to increase the contrast of the markers for better readability. For some applications, we also mounted a projector (Epson VS250) 100 cm above the table to present graphical information.
As the system tracks the robots from below, the projected image does not affect tracking.
Figure~\ref{fig:system-tracking-setup} illustrates our current setup.

\subsection{Control System}
The control algorithm is as follows. First, the user specifies the target position, orientation, and length of each actuator. Then, given the position of each robot ($R_1, \cdots, R_N$) and the target points ($T_1, \cdots, T_M$), the system calculates the distance between each position and constructs a matrix $M$ ($M_{i, j} = $ distance between $R_i$ and $T_j$). Given this distance matrix $M$, the system solves the target assignment problem using the Munkres assignment algorithm. Once a target position is assigned to each robot, the main computer continuously sends commands to direct individual robots to their respective target locations. To avoid collisions, we used the Reciprocal Velocity Obstacles (RVO) algorithm~\cite{van2008reciprocal}. This algorithm provides an incremental target vector from the current position for each robot. Based on the next target position, we compute the PWM signal for four pins of two wheels (i.e., left wheel: forward or backward, right wheel: forward or backward). We direct and navigate each robot using a proportional integral derivative (PID) control. Once the robot reaches its target position, it rotates to face the desired direction. The accuracy threshold of position and orientation are 1 cm and 5 degrees respectively. Once located and oriented to the target position and orientation within this threshold, the robot stops moving and extends its actuator to construct a shape. When the robot is assigned a new target position, it retracts the actuator, then repeats the same process. 

\begin{figure}[!h]
\vspace{-0.2cm}
\centering
\includegraphics[width=3in]{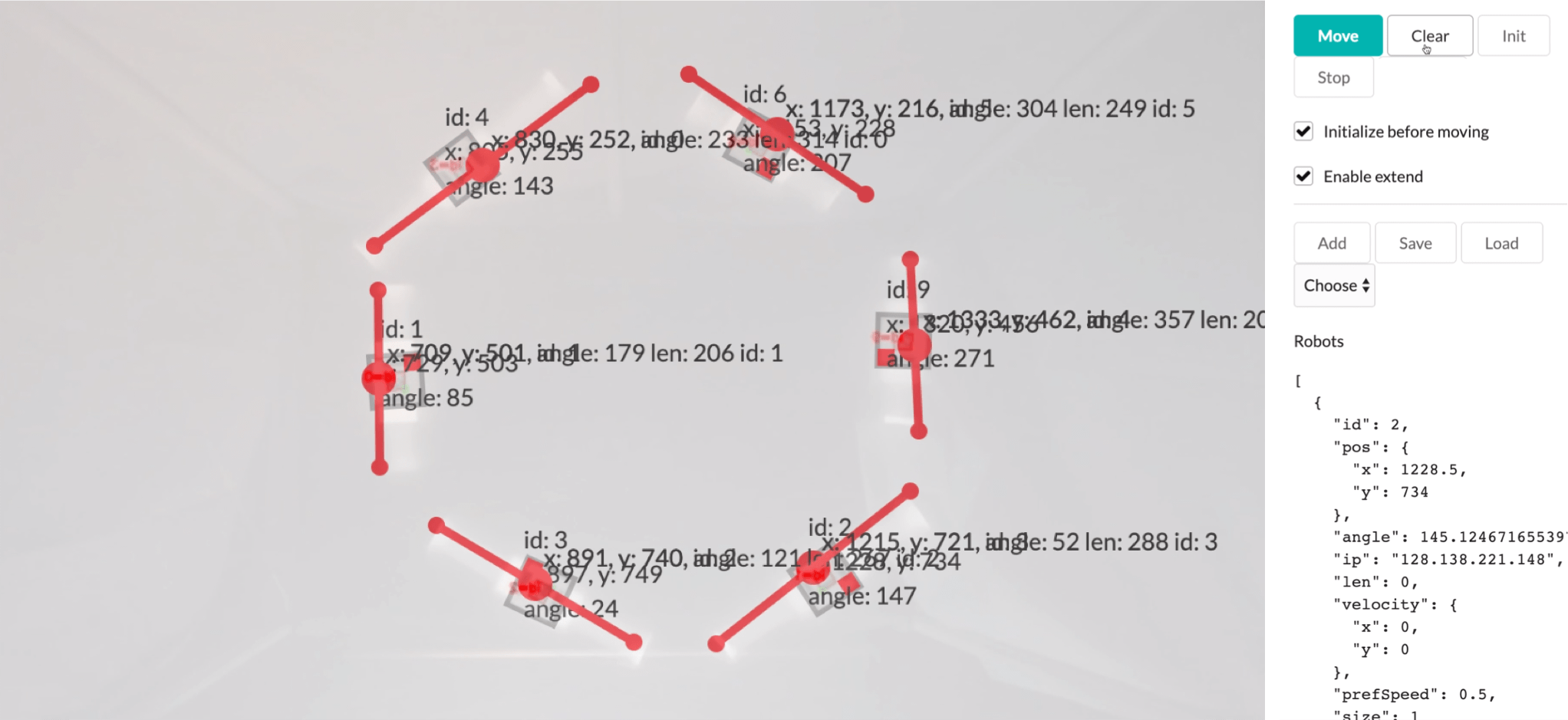}
\caption{Web-based interface to create a target shape and keyframe animation based on a user's drawing or SVG image.}
~\label{fig:system-control}
\vspace{-0.6cm}
\end{figure}

To enable the user to easily specify a target shape, we created a web-based interface where users draw a shape or upload an SVG image (Figure~\ref{fig:system-control}). The user draws a set of lines, then the main computer calculates target positions, orientations, and actuator lengths to start sending commands. The user can create a keyframe animation by drawing a sequence of frames. The user can also upload an SVG image: the system first computes a contour of the image, simplifies the contour into lines given the number of robots, drives each robot to the center of each line, and extends the robot's arms to display the line.

\subsection{Interaction Capability}
\begin{figure}[!h]
\vspace{-0.2cm}
\centering
\includegraphics[width=3.4in]{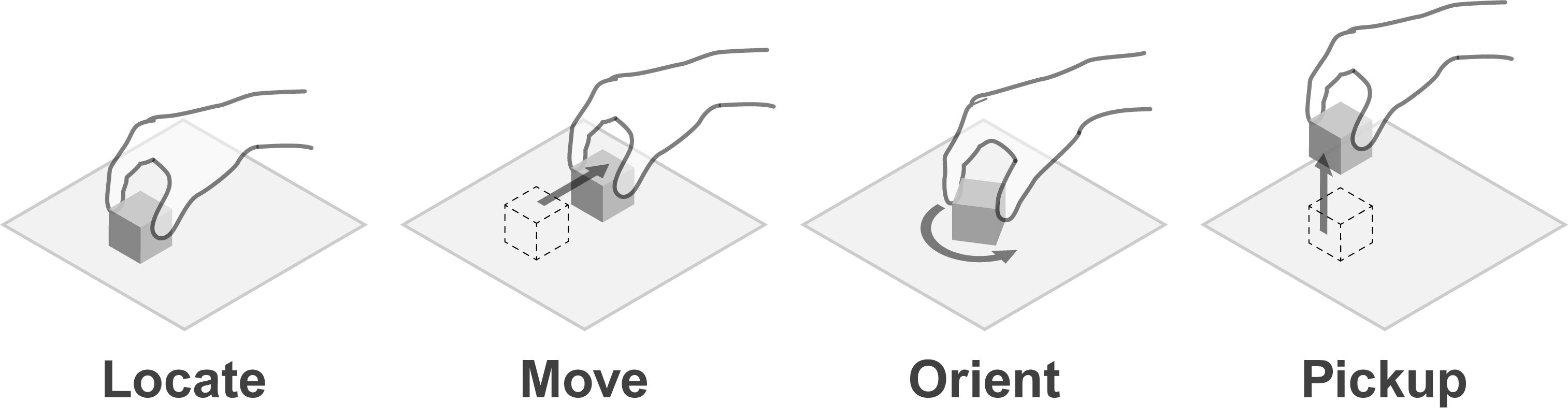}
\caption{Interaction capability of ShapeBots.}
~\label{fig:system-inputs}
\vspace{-0.6cm}
\end{figure}

We can use the same mechanism to track user input. Figure~\ref{fig:system-inputs} illustrates four different types of user interaction that our system supports: place, move, orient, and pick-up. The system recognizes as user inputs movement or rotation of a marker that it did not generate. When the system detects a new marker or loses an existing marker, it recognizes that the user is placing or picking up a robot.
\changes{
Robots that the systems are driving are not candidates for user input. Thus, when unintended events occur (e.g., bumping or colliding), the system distinguishes these from user input, as it is driving the robots.
}
A single user can manipulate multiple robots with two hands, or multiple users can interact with the robots. In addition, by leveraging OpenCV's object detection algorithm, the system can detect the position and shape of objects on the table, such as pens, a sheet of paper, coffee cups, and phones.

\subsection{Technical Evaluation}
We evaluated the technical capabilities of ShapeBots, including: 
{\bf 1)} speed of the robot and linear actuators, 
{\bf 2)} accuracy of extending the linear actuator with open-loop control, 
{\bf 3)} position and orientation accuracy of the robot with our tracking and control scheme, 
{\bf 4)} strength of the robot when moving external objects,
{\bf 5)} load-bearing limit of the linear actuator,
\changes{{\bf 6)} latency of a control loop, and {\bf 7)} robustness of tracking.}
We also developed a simulator to compare the rendering capabilities of ShapeBots with non self-transformable swarm robots.

\subsubsection{Method}
{\bf 1)} We measured the speed of the robot and linear actuator via video recordings of the robot moving next to a ruler. 
{\bf 2)} For the extension accuracy of the linear actuator, we took three measurements of the extension length given the same motor activation duration for three different duration values. 
{\bf 3)} For position and orientation accuracy, we logged the distance and angle deviation between the robot and its target with our control system. 
{\bf 4)} We employed Vernier's dual-range force sensor (accurate to 0.01N) for force measurements. For the pushing force, we measured the maximum impact force against the sensor.
{\bf 5)} For the load-bearing limit, we gradually increased the load on the end-cap of the linear actuator and measured the force just as the reel buckled.
\changes{{\bf 6)} We measured the latency of each step (i.e., Wi-fi communication, tracking, and computation of path planning) by comparing timestamps from the start of each event to its conclusion with 10 attempts for 10 robots.}
\changes{{\bf 7)} For the robustness of tracking, we evaluated the error rate by measuring the frequency of lost marker tracking during movement.}

\subsubsection{Results}

\begin{figure}[!h]
\vspace{-0.2cm}
\centering
\includegraphics[width=3.4in]{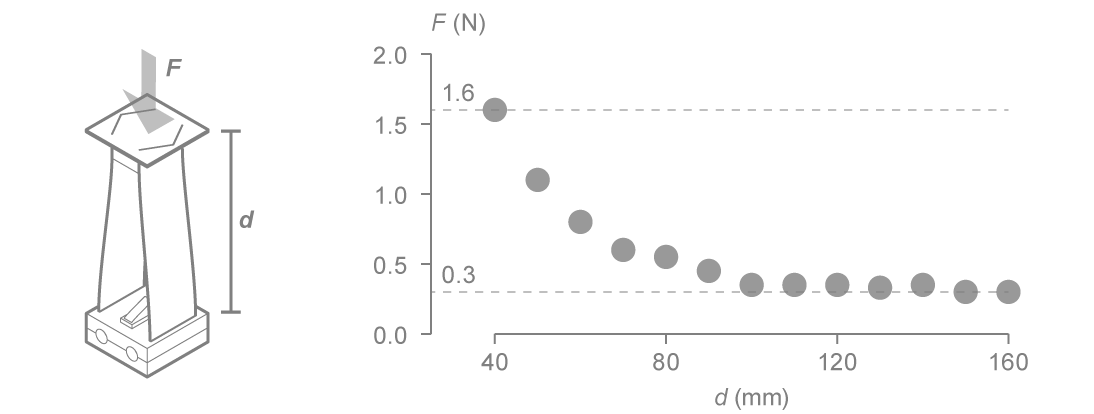}
\caption{Load-bearing limit of linear actuator at different lengths.}
~\label{fig:forcegraph}
\vspace{-0.6cm}
\end{figure}

The table summarizes the results of measurements for each of {\bf (1 - 3)} 
{\bf 4)} We found that the moving robot can deliver a maximum force of 0.24N. An attached linear actuator can deliver the same force up to 130mm extension; beyond that, we observed a significant decrease (approximately 0.9N). 
{\bf 5)} We observed diminishing vertical load-bearing capability for the linear actuator as extension length increases, plateauing at about 0.3N beyond 100mm (Figure~\ref{fig:forcegraph}).
\changes{{\bf 6)} The measured average latency was 68 ms (Wi-fi communication), 39 ms (tracking), 44 ms (computation of path planning), which takes 151 ms for the total latency of the control loop with 10 robots.}
\changes{{\bf 7)} With 10 attempts, the system loses tracking once every 1.9 sec on average. Given 151 ms for one control loop, the error rate was 7.9\%. To decrease the false positive user input detection, the system distinguishes these errors from user input by setting a threshold.}

\setlength{\tabcolsep}{6pt}
\renewcommand{\arraystretch}{1.1}
\begin{tabular}{ |l|l| } 
 \hline
 Robot Maximum Speed & 170 mm/s \\ 
 Linear Actuator Speed & 33 mm/s \\ 
 Average Linear Actuator Extension Error & 3 mm \\
 Average Position Error & 3.2 mm \\
 Average Orientation Error & 1.7 deg \\
 Latency & 151 ms \\
 Tracking Error Rate & 7.9\% \\
 \hline
\end{tabular}

\begin{figure}[!h]
\centering
\includegraphics[width=3.4in]{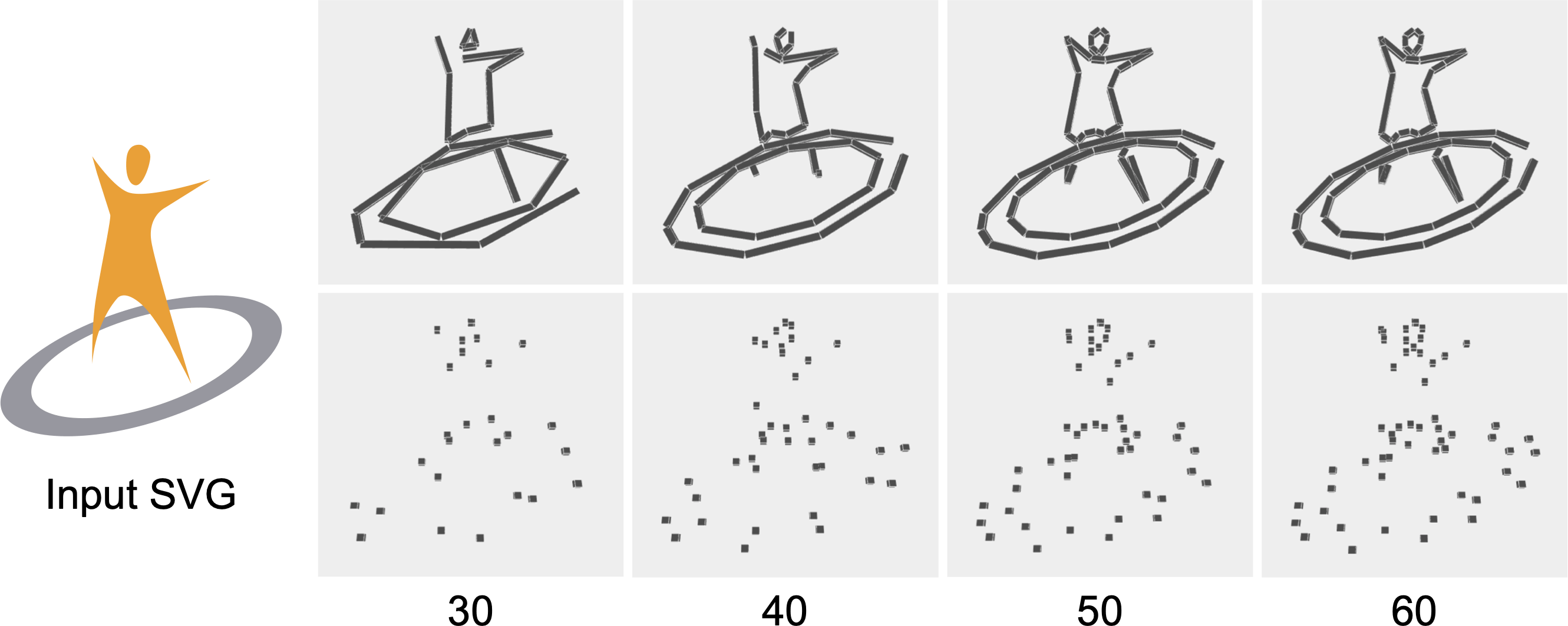}
\caption{Simulation results comparing ShapeBots (top) with swarm robots (bottom). Left to right: original SVG image, rendering simulation results with 30, 40, 50, and 60 robots respectively.}
~\label{fig:system-simulation}
\vspace{-0.6cm}
\end{figure}

Figure~\ref{fig:system-simulation} highlights the advantage of ShapeBots for rendering contours compared to non self-transformable swarm robots. Using a software simulation, we demonstrate how ShapeBots renders an SVG input at different swarm sizes.
\changes{
Our simulation algorithm has three steps: 1) Convert an input SVG path into a list of polylines, 2) Set the vertex of each polyline as target location (non-transform swarm), or the midpoint of the line (ShapeBots), 3) Assign the goal and move, orient, and transform the robot. 
We localize the target position {\it to draw a contour} of the SVG image, instead of {\it filling the image} as seen in existing algorithm~\cite{alonso2012image} because it would be a better comparison for the line-based swarm (dot vs line).
The interactive simulator and its source code are available online~\footnote{\url{https://ryosuzuki.github.io/shapebots-simulator/}}.
}

\section{Application Scenarios}

\subsection{Interactive Physical Displays and Data Representations}

\subsubsection{Interactive Data Physicalization}
\begin{figure}[!h]
\vspace{-0.2cm}
\centering
\includegraphics[width=3.4in]{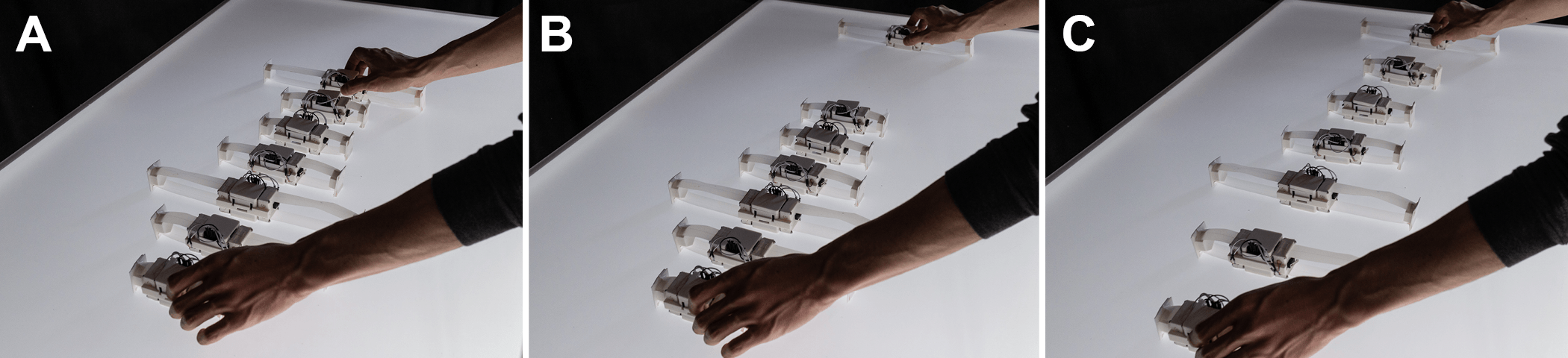}
\caption{An interactive and animated sine wave. A) Animated sine wave. B) When the user moves one element, C) then each robot can collectively move to change the spatial period of the wave.}
~\label{fig:applications-sine-wave}
\vspace{-0.6cm}
\end{figure}

One interesting application area is interactive data physicalization~\cite{jansen2015opportunities, taylor2015data}. For example, in Figure~\ref{fig:applications-sine-wave} seven robots transform individually to represent a sine wave. These representations are interactive with user inputs: when the user moves the end robot to the right, the others move to change the wavelength. The user can interactively change the amplitude of the wave by specifying the maximum length. 

\begin{figure}[!h]
\vspace{-0.2cm}
\centering
\includegraphics[width=3.4in]{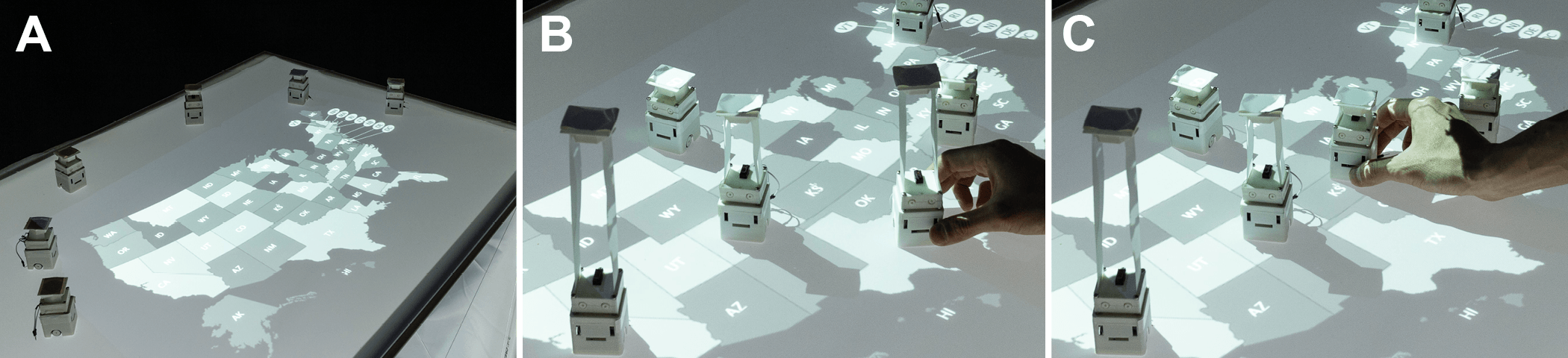}
\caption{Embedded data physicalization on a map. A) Projected US map. B) When the user selects the dataset, the ShapeBots move to position and visualize data with their heights. C) When moved, the robots change their heights accordinly.}
~\label{fig:applications-map}
\vspace{-0.6cm}
\end{figure}

ShapeBots also support transforming data into different representations such as bar graphs, line charts, and star graphs. The user can place and move robots, which enables embedded data representations~\cite{willett2017embedded}. For example, ShapeBots on the USA map physicalize map data; each robot changes its height to show the population of the state it is on (Figure~\ref{fig:applications-map}). Users can interact with the dataset by placing a new robot or moving a robot to a different state, and the robots update their physical forms to represent the respective population. 

Other examples of distributed representations include showing the magnitude and orientation of wind on a weather map, or physicalizing magnetic force fields. This physical data representation could be particularly useful for people with visual impairments~\cite{guinness2018haptic, suzuki2017fluxmarker}.

\subsubsection{Interactive Physical Display}
\begin{figure}[!h]
\centering
\includegraphics[width=3.4in]{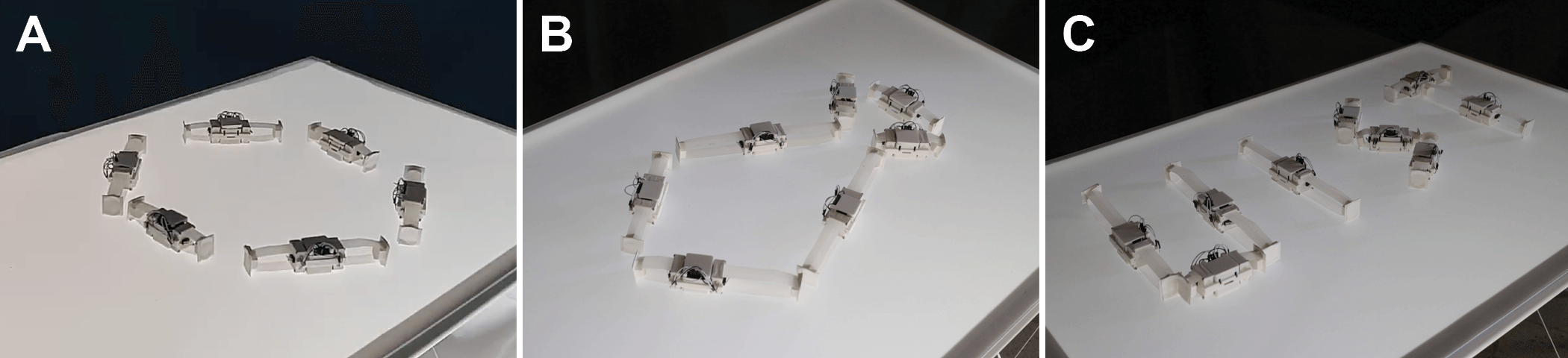}
\caption{Example shapes. A) hexagon, B) fish, C) text.}
~\label{fig:applications-display}
\vspace{-0.6cm}
\end{figure}

\begin{figure}[!h]
\centering
\includegraphics[width=3.4in]{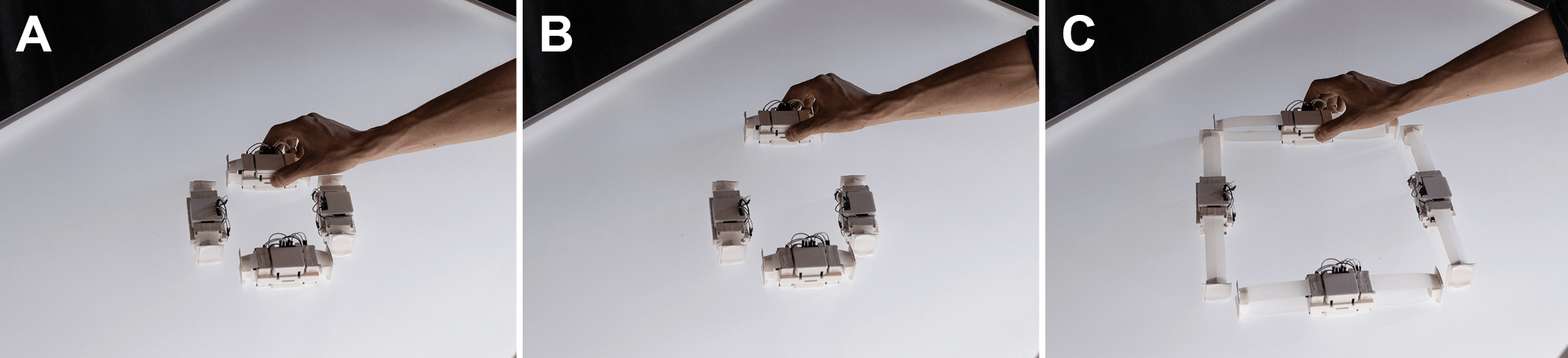}
\caption{Interactive shape change. A) A small rectangle shape. B) If the user moves one element, C) then the robots change positions and lengths to scale the square.}
~\label{fig:applications-interaction}
\vspace{-0.6cm}
\end{figure}

ShapeBots can also act as an interactive physical display.  Figure~\ref{fig:applications-display} shows how ShapeBots can render different shapes. 
It also allows users to preview a shape. For instance, when reading a picture book of animals, children can visualize the fish with ShapeBots at actual size (Figure~\ref{fig:applications-display}B). 
Figure~\ref{fig:applications-interaction} demonstrates the input and output capabilities as an interactive tangible display. Four robots first move to display a small rectangle. When the user moves a robot, the others change positions and lengths to scale the shape. The user can also move robots to rotate or translate the shape. 

\begin{figure}[!h]
\centering
\includegraphics[width=3.4in]{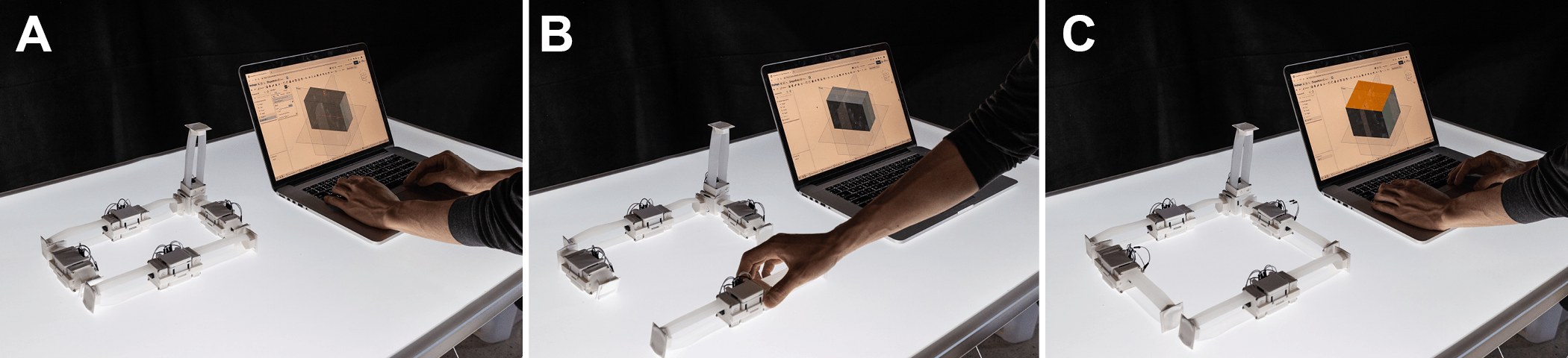}
\caption{Physical preview for the CAD design. A) ShapeBots provide a physical preview synchronized with the computer screen. B) When the user manipulates the element, C) then it updates the digital design of the CAD software.}
~\label{fig:applications-cad}
\vspace{-0.6cm}
\end{figure}

Similarly, ShapeBots can provide a physical preview of a CAD design. Figure~\ref{fig:applications-cad} shows a user designing a box. ShapeBots physicalizes the actual size of the box. The design and physical rendering are tightly coupled; as the user changes the height of the box in CAD software, the ShapeBots change heights accordingly (Figure~\ref{fig:applications-cad}A). The user can change the parameters of the design by moving robots in the physical space, and these changes are reflected in the CAD design (Figure~\ref{fig:applications-cad}B-C). 

\subsection{Object Actuation and Physical Constraints}

\subsubsection{Clean up Robots}
\begin{figure}[!h]
\vspace{-0.2cm}
\centering
\includegraphics[width=3.4in]{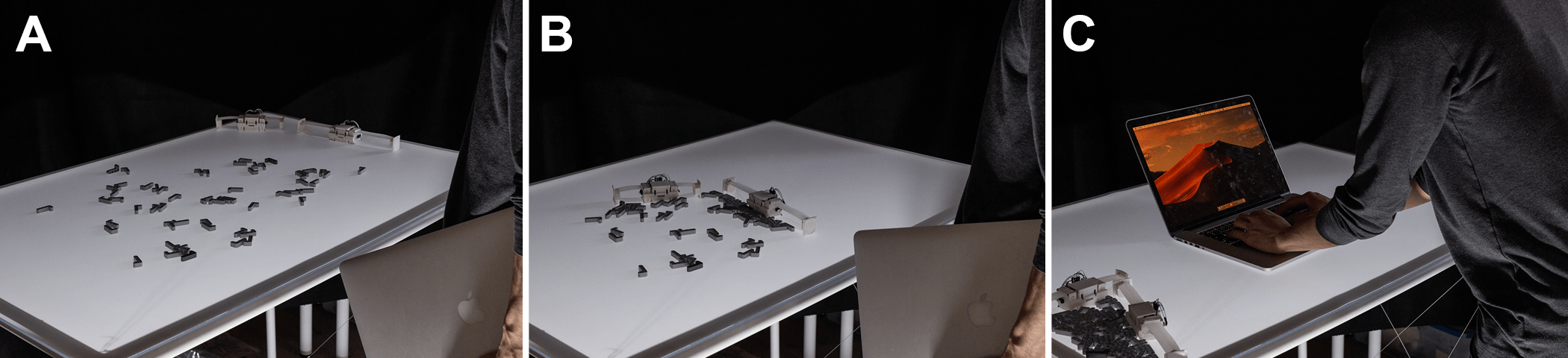}
\caption{Clean up robots. A) A desk is filled with debris. B) Two robots starts moving and wiping the debris. C) Once the robots finish cleaning up, the user can start using the workspace.}
~\label{fig:applications-clean-up}
\vspace{-0.6cm}
\end{figure}

Another practical aspect of ShapeBots is the ability to actuate objects and act as physical constraints. As an example, Figure~\ref{fig:applications-clean-up} shows two robots extending their linear actuators to wipe debris off a table, clearing a workspace for the user. 

\subsubsection{Tangible Game}
\begin{figure}[!h]
\vspace{-0.2cm}
\centering
\includegraphics[width=3.4in]{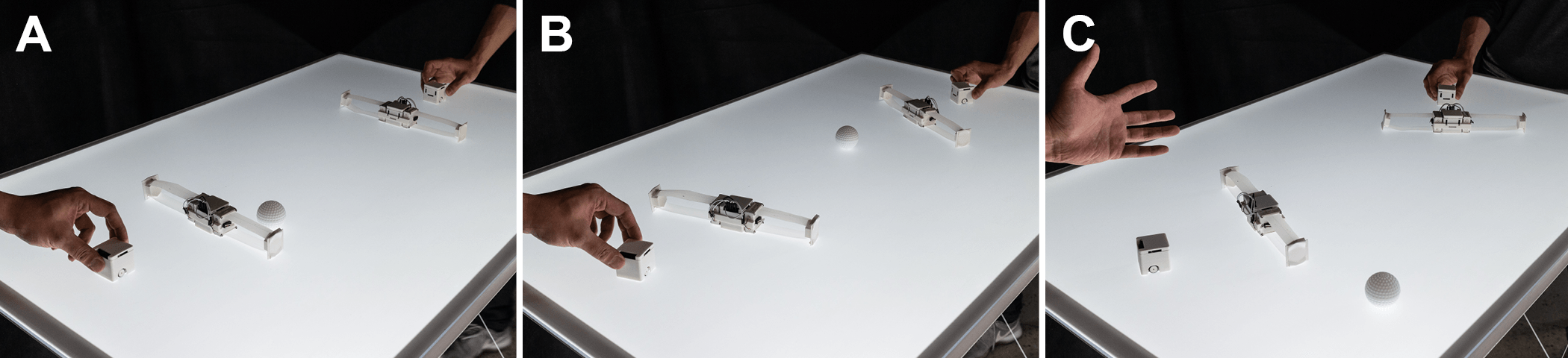}
\caption{Tangible game controllers. A) Two user start playing a table football game. B) Each user uses one ShapeBot as a controller and another ShapeBot to hit the ball. C) The user can play with these ShapeBots like table football.}
~\label{fig:applications-tangible-game}
\vspace{-0.8cm}
\end{figure}

ShapeBots can be employed as a tangible gaming platform. Figure~\ref{fig:applications-tangible-game} illustrates two users playing a table football game using two extended robots. The user controls a robot acting as the pinball arms whose position and angle are synchronized to a controller robot. Users hit or block a ball to target a goal, similar to table football.

\subsection{Distributed Dynamic Physical Affordances}

\subsubsection{In-situ Assistants}
\begin{figure}[!h]
\vspace{-0.2cm}
\centering\includegraphics[width=3.4in]{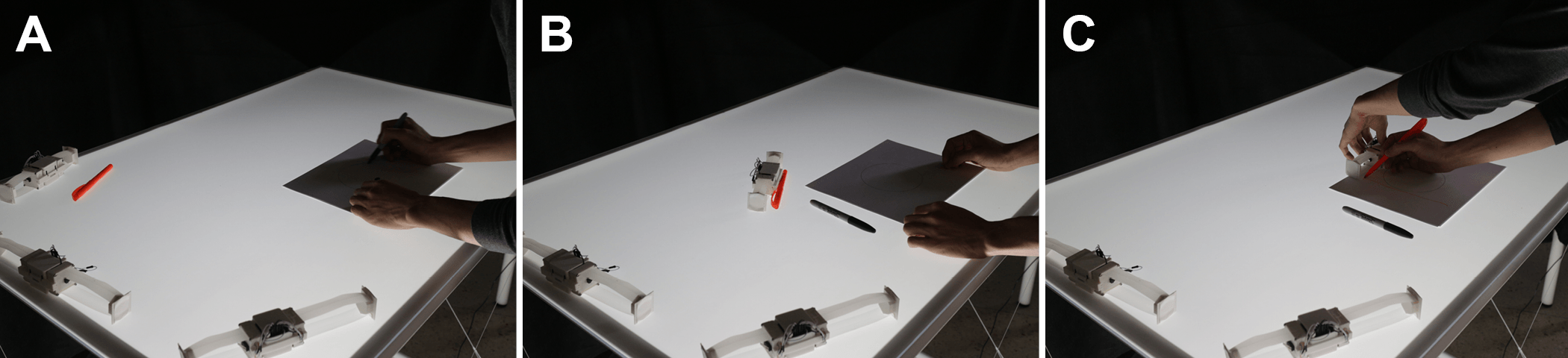}
\caption{Shape-changing Tools. A) A user is working on a desk. B) When the user needs a pen, ShapeBots can bring it. C) ShapeBots can be also used as a tool like ruler.}
~\label{fig:applications-tools}
\vspace{-0.6cm}
\end{figure}

In another scenario, the ShapeBots brings tools to the user. For instance, when the user needs a pen, a robot extends its actuators and pushes the pen to the user (Figure~\ref{fig:applications-tools}A-B). These robots can also be used as tools. In the same example, when the user needs to draw a line with a certain length, the user specifies the length, then the robot extends its length to serve as a ruler (Figure~\ref{fig:applications-tools}C). The user can also bend the linear actuator or using multiple robots to draw a curved line or other shapes. 

\subsubsection{Dynamic Fence}
\begin{figure}[!h]
\vspace{-0.2cm}
\centering
\includegraphics[width=3.4in]{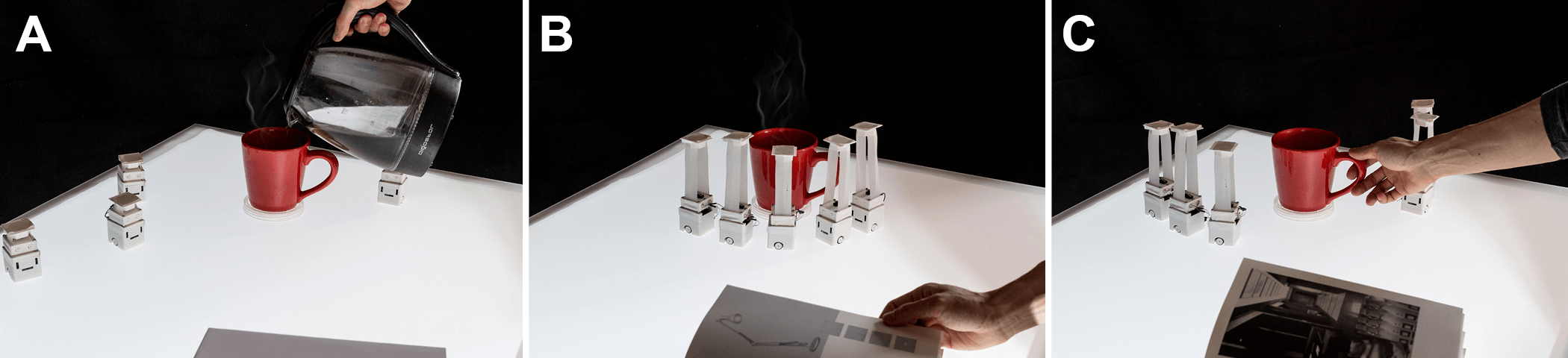}
\caption{Distributed dynamic physical affordances. A) A user pours hot coffee. B) ShapeBots create a vertical fence to prevent the user from grabbing the cup. C) Once the coffee has cooled, the ShapeBots disperse.}
~\label{fig:applications-fence}
\vspace{-0.8cm}
\end{figure}

By leveraging the capability of locomotion and height change of each robot, ShapeBots can create a dynamic fence to hide or encompass existing objects for affordances. 
For example, Figure~\ref{fig:applications-fence} illustrates this scenario. When the user pours hot coffee into a  cup, the robots surround the cup and change their heights to create a vertical fence.
The vertical fence visually and physically provides the affordance to indicate that the coffee is too hot and not ready to drink. Once it is ready, the robots start dispersing and allow the user to grab it. These scenarios illustrate how the distributed shape-changing robots can provide a new type of affordance, which we call distributed dynamic physical affordances.

\section{Discussion and Design Space}
\begin{figure}[!t]
\centering
\includegraphics[width=3.4in]{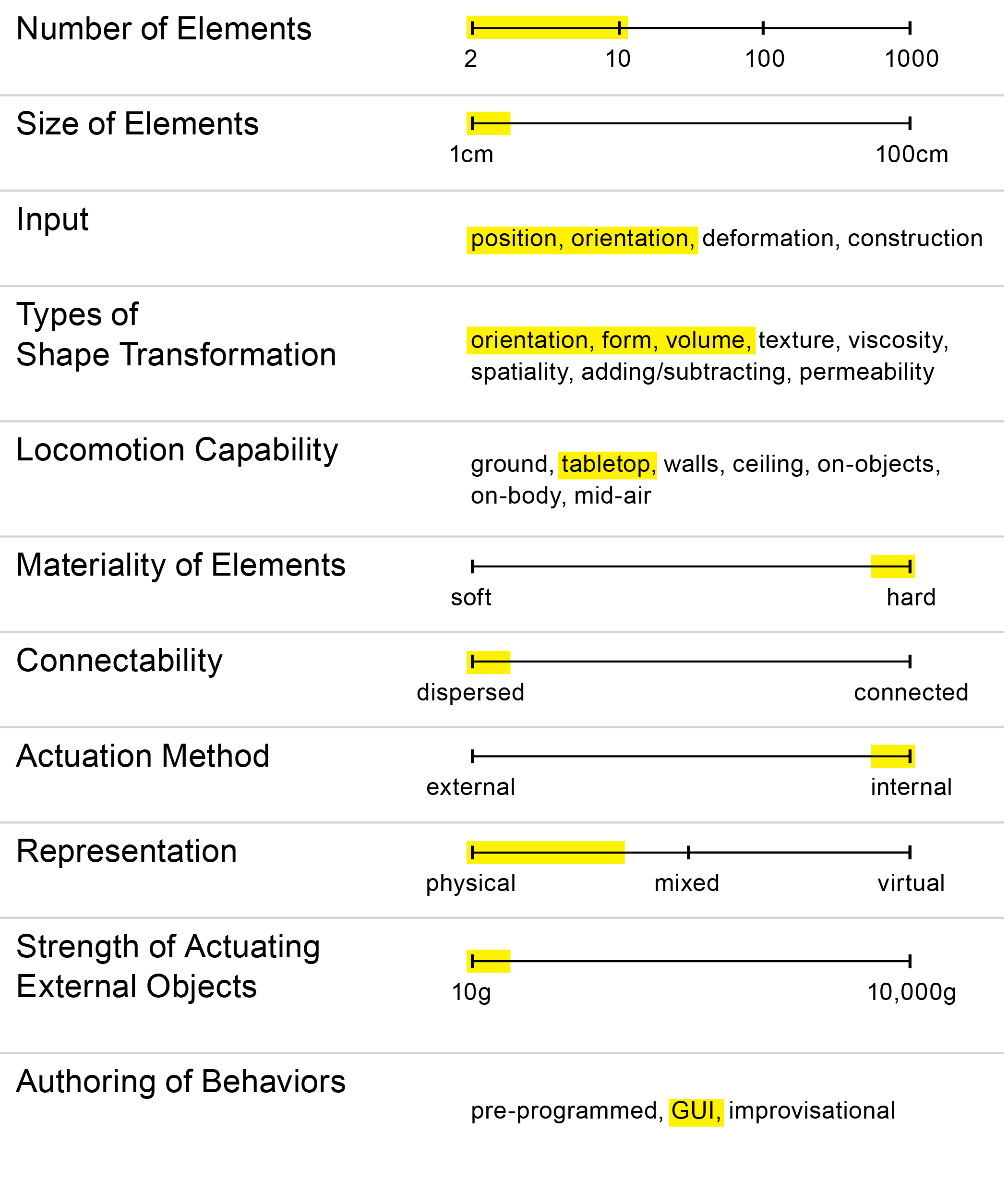}
\caption{Design space of shape-changing swarm user interfaces.}
~\label{fig:design-space}
\vspace{-0.8cm}
\end{figure}

This section explores the broader design space of shape-changing swarm user interfaces and discusses how future work can address open research areas. The table identifies dimensions of the design space of shape-changing swarm user interfaces (Figure~\ref{fig:design-space}). The highlighted regions represent where ShapeBots fit within the design space.

\subsubsection{Number of Elements}
The number of elements is a key design dimension of shape-changing swarm interfaces. The more elements, the more expressive and detailed rendering is possible. However, the number of elements and the shape-changing capability of each element are complementary. By increasing degrees of freedom of shape-change, sparsely distributed robots can better fill gaps between them. 
Moreover, larger numbers of elements create interesting interaction challenges: how does the user manipulate 10 or even 100 elements at once? For example, a user could manipulate multiple elements by grouping, creating constraints between multiple objects, and/or introducing higher-level abstraction. 

\subsubsection{Size of Elements}
\begin{figure*}[!t]
\centering
\includegraphics[width=1\textwidth]{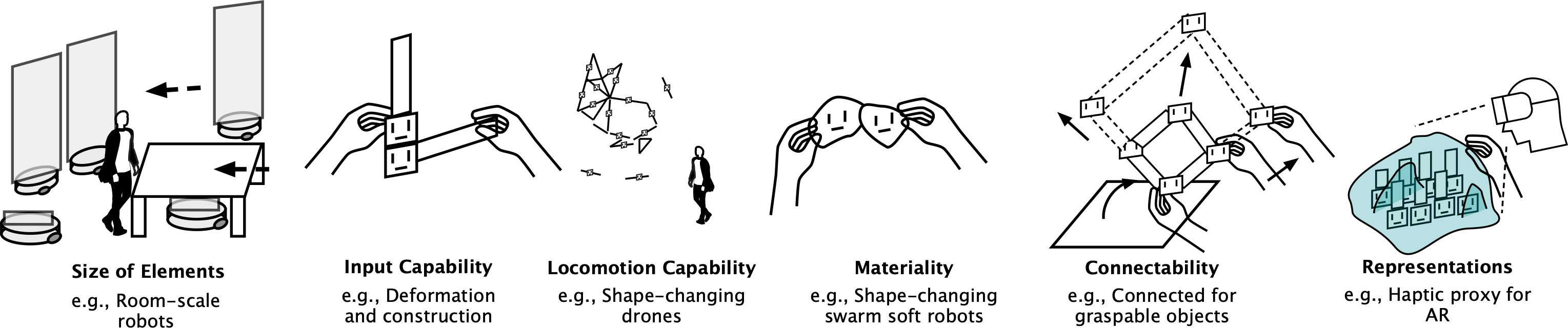}
\caption{Design space of shape-changing swarm user interfaces illustrating future research opportunities.}
~\label{fig:future-opportunities}
\vspace{-0.8cm}
\end{figure*}

The size of the swarm elements is another dimension. This paper focuses on small (3 - 4 cm) robots, but other sizes would open up new application domains. For example, room-size shape-changing swarm robots could produce dynamic furniture (e.g., transforming chairs and tables) or modify the spatial layout of a room through movable and expandable walls (Figure~\ref{fig:future-opportunities}A). Today, most large robots are seen in factories and warehouses, but as they enter everyday environments (e.g., cleaning or food delivery robots), we see opportunities for shape-changing robots to enable people to interact with and reconfigure their environments~\cite{knight2017get, spadafora2016designing}.

Alternatively, each swarm element could be further reduced in size. Le Goc et al.~\cite{le2016zooids} introduced a continuum between ``things'' and ``stuff'' to discuss this aspect. At this resolution, one might assume that the shape-changing capability of each swarm element is not required. But many materials found in nature---from living cells (e.g., chloroplast) to molecules (e.g., polyisoprene)---use self-transformation to change the overall structure or physical properties (e.g., stiffness, form). In the same way, shape-changing swarms of {\it stuff} can exhibit overall dynamic material properties or structural change resulting from the aggregation of individual small shape changes. 

\subsubsection{Input}
Our current prototype supports four types of user inputs: place, move, orient, and pick up. Supporting other types of inputs can enhance richer interaction possibilities (Figure~\ref{fig:future-opportunities}B). For example, if the user can also manually extend lines or deform the shape of robots, new types of applications could be possible, such as using these robots as a physical tape measure like SPATA~\cite{weichel2015spata} or to demonstrate shape changes through tangible demonstration like Topobo~\cite{raffle2004topobo}.

\subsubsection{Types of Shape Transformation}
The types of shape change of the swarm elements affect the expressiveness and affordances of the interface. Rasmussen et al.~\cite{rasmussen2012shape} classified eight different types of shape change: orientation, form, volume, texture, viscosity, spatiality, adding/subtracting, and permeability. Of these, the current ShapeBots implementation supports form, volume, and orientation change. Additional transformations could enrich the capability of swarm user interfaces. For example, each robot could change its texture to communicate its internal state, similar to~\cite{hu2018soft}. Another interesting aspect of shape-changing swarms is the potential to simultaneously achieve both topologically equivalent (e.g., form, volume) and non-topologically equivalent shapes (e.g., adding/subtracting, permeability) by combining individual and collective shape transformations. This capability could expand the taxonomy and design space of shape transformation.

\subsubsection{Locomotion Capability}
Camera-based tracking limits the locomotion capability of Shapebots to a tabletop surface. However, future shape-changing swarm robots could also cover walls, ceilings, objects, the user's body and hover in mid-air. Swarm robots on walls, windows, and building facades could serve as an expressive tangible public information display. A swarm of ceiling crawling self-transforming robots could pick up objects and move them from one place to another through extendable arms. Using a sucking mechanism demonstrated in Skinbots~\cite{dementyev2018epidermal}, robots could move on-body or on-object. Mid-air drone swarms~\cite{braley2018griddrones}, with added shape-changing capabilities could form more expressive mid-air displays like vector graphics or a 3D mesh structure (Figure~\ref{fig:future-opportunities}C).

\subsubsection{Materiality of Elements}
Most swarm robots, including our current prototype, are made of rigid materials. However, soft swarm robots made of malleable materials could exhibit other expressive self-transformation capabilities such as changes in volume~\cite{niiyama2014weight} and stiffness~\cite{follmer2012jamming}. Moreover, an increasing number of works investigate modular soft robots~\cite{nakayama2019morphio, robertson2017new}. For example, soft modular robotic cubes~\cite{vergara2017soft} demonstrated that deformation of modules can be used for locomotion. Although fitting actuation into the size of the current swarm robot is a technical challenge, a tiny pump that fits into a small body could address this. From the interaction perspective, it would be interesting if users could construct an object using these soft robots like pieces of clay, and the constructed soft object can transform itself to another shape (Figure~\ref{fig:future-opportunities}D). 

\subsubsection{Connectability}
In the ShapeBots prototype, the swarm elements do not physically connect, but the ability of swarm robots to connect to each other would enable graspable 3D shapes. For example, the connection between lines enables the user to pick up a rendered object while the object can dynamically change its shape or scale in the user's hand (Figure~\ref{fig:future-opportunities}E). With a sufficient number of lines, such objects can represent arbitrary shapes, similar to LineFORM~\cite{nakagaki2015lineform}. By enabling vertically extendable robots to connect, one can achieve a handheld shape display that can dynamically reconfigure a 3D shape. These objects are particularly useful to provide a haptic proxy for virtual objects~\cite{zhao2017robotic} or an instant preview of a 3D design~\cite{suzuki2018dynablock}.

\subsubsection{Actuation Method}
Our prototype contains all components (locomotion, actuation, and battery) within each robot, but further reducing the size to less than 1 cm poses a technical challenge. External actuation methods might address some of these problems. For example,~\cite{patten2007mechanical, pelrine2012diamagnetically, strasnick2017shiftio, suzuki2018reactile} demonstrated the use of electromagnetic coil arrays to actuate swarm of small magnets (e.g., 3 mm - 1 cm). External actuation could enable locomotion and transformation of swarm objects, similar to Madgets~\cite{weiss2010madgets}.

\subsubsection{Representation}
The physical content representation of shape-changing swarm robots can be combined with other modalities. As we demonstrated, with projection mapping, graphics can present information that is difficult to convey solely through physical representations. Alternatively, graphics can be shown through internal LED lights (e.g., represent groups in data plots through color). An interesting research direction would leverage the physicality of robots to provide haptic feedback for virtual or augmented reality (Figure~\ref{fig:future-opportunities}F). For example, we demonstrated a scenario where the robots show a physical preview of CAD design. With mixed reality the user can overlay information on top of the robots like Sublimate~\cite{leithinger2013sublimate} or MixFab~\cite{weichel2014mixfab}.

\subsubsection{Strength of Actuation}
The strength and stability of the actuation mechanism is another key design dimension. Our linear actuator can only move lightweight objects and is not stable enough to actuate or withstand heavy objects. Projects like G-Raff~\cite{kim2015g} demonstrated a similar design in a larger form factor with a more stable reel and stronger motor to lift heavier objects such as phones. With stronger actuators, swarm robots could act as adaptable physical objects e.g., it can sometimes be an adjustable bookstand, and sometimes be a footrest. Stronger actuation is particularly interesting for larger swarm robots. Room-size shape-changing robots that can lift and carry heavy furniture and can dynamically transform a room layout.

\subsubsection{Authoring of Behaviors}
The final design dimension is how to author and program the shape-changing swarm behavior. Currently, the user programs ShapeBots through a GUI based authoring tool. But, the authoring capability of animation or interactive applications is still primitive (e.g., using keyframe animation or script based programming). It is important to allow users and designers to improvisationally author behaviors within a physical space. One promising approach is to adapt programming by demonstration, but it is still challenging to abstract these demonstrations and create interactive applications for large swarms. Reactile~\cite{suzuki2018reactile} explored tangible swarm UI programming by leveraging direct manipulations. Future research will investigate how these practices can be applied to shape-changing swarm user interfaces.

\section{Limitations and Future Work}
The previous section discusses high-level research directions that go beyond the scope of ShapeBots. Here, we describe the specific limitations and future work for our current design and implementation. 

\textit{\changes{Connection}}:
The current design uses wires and pin headers to electrically connect actuators to each robot. To make it easier to detach and reconfigure, we will consider a magnetic connection and pogo pins, similar to LittleBits~\cite{bdeir2009electronics}. 
\changes{In addition, we are interested in exploring connection mechanisms to assemble these robots into graspable 3D shapes.}

\textit{\changes{Scalability}}:
\changes{In our prototype, we tested our concept with twelve robots. When we try to scale to hundreds instead of tens, we will face a number of technical challenges. For example, our tracking relies on the ArUco algorithm~\cite{garrido2014automatic, romero2018speeded}.
Based on their evaluation, the mean time of detection is 1.4 ms in 1080p, which can maintain the current frame rate with 10-20 robots, but for over 100 robots, this latency may cause problems.
Moreover, in a practical point of view, the more significant problem is the robots will bump into one another, given the fixed driving area. Compared to non-transformable swarms, this becomes more critical. This is why the robot needs to retract the reel before moving, although it causes a transition problem in animation. A possible solution for future work is to only retract when path planning indicates a collision.}

\textit{\changes{Low Force Output}}: 
\changes{
As we discussed in technical evaluation section, the force for movement and linear actuation is relatively weak for applications such as haptics or heavy object actuation.
The driving force can be improved by increasing friction of wheel, similar techniques used in~\cite{kim2019swarmhaptics}. 
To improve the structural stability, we expect different materials for reel or using different mechanisms (e.g., scissor structure) may increase the output force.
}

\textit{\changes{Input Modalities}}:
\changes{In this paper, we primarily focused on output, with less emphasis on rich input capabilities like touch inputs or manual reel extraction. For future work, it is interesting to investigate how the user can interact with robots in improvisational ways to construct and deform the shape.}

\textit{\changes{More Complex Geometries}}:
\changes{
Application scenarios with more complex geometries (e.g., using area/volumetric expansion for dynamic bubble chart) were difficult to demonstrate due to resolution and size constraints.
Also, currently, each robot can only control two individual actuators. Curvature control is done by simply changing the length of the two strips. Having more individual actuator controls could support more expressive transformation such as transforming from dot to line, triangle, and diamond with a single robot. 
However, it also introduces a new design consideration.
For example, a tradeoff between ``fewer and more complex robots that support expressive geometries'' vs ``a higher number of simpler robots'' is particularly unique design consideration for shape-changing swarm robots.
}

\textit{\changes{Battery Charging}}:
Our robots require recharging the battery every hour. To address this issue, we will investigate wireless power charging~\cite{uno2018luciola} or other continuous power supply methods~\cite{klingner2014stick}. 

\textit{\changes{Different Tracking Mechanism}}:
We are also interested in tracking mechanisms that cover larger areas and work outside of a specifically designed table. We will investigate laser-based tracking~\cite{cassinelli2005smart} for this purpose. 

\textit{\changes{User Evaluation}}: 
For future work, we will conduct a formal user evaluation of our prototype. In particular, we are interested in evaluating the benefits and limitations of the current approach for the specific applications of data physicalization and accessibility assistance. 

\section{Conclusion}
This paper proposed \emph{shape-changing swarm robots}, a new approach to distributed shape-changing interfaces that form shapes through collective behaviors of self-transformable robots. We demonstrated this approach through ShapeBots. Our novel miniature linear actuators enable each small robot to change its own shape. In a swarm, they can achieve various types of shape representation and transformations. A set of application scenarios show how these robots can enhance the interaction and affordances of shape-changing interfaces.
We hope this work provokes a discussion on the potential of shape-changing swarm user interfaces towards the future vision of ubiquitous and distributed shape-changing interfaces.

\section{Acknowledgements}
We thank Takayuki Hirai and Shohei Aoki for their help of initial hardware implementation.
This research was supported by the JST ERATO Grant Number JPMJER1501, NSF CAREER award IIS 1453771, and the Nakajima Foundation.

\balance
\bibliographystyle{SIGCHI-Reference-Format}
\bibliography{references}

\end{document}